\documentclass{article}

\usepackage{microtype}
\usepackage{graphicx}
\usepackage{booktabs} %

\usepackage{hyperref}

\usepackage{url}
\usepackage[utf8]{inputenc}
\usepackage{booktabs} %
\usepackage{amsfonts} %
\usepackage{nicefrac} %
\usepackage{microtype} %
\usepackage{graphicx}
\usepackage{natbib}
\usepackage{gensymb}
\usepackage{color}
\usepackage{caption}
\usepackage{subcaption}
\usepackage{wrapfig}
\usepackage{soul}
\usepackage[ruled,longend, algo2e]{algorithm2e}
\usepackage[textwidth=1.2in,textsize=tiny]{todonotes}
\usepackage{diagbox}
\usepackage{hhline}
\usepackage{multirow}

\usepackage[accepted]{icml2024}

\usepackage{amsmath}
\usepackage{amssymb}
\usepackage{mathtools}
\usepackage{amsthm}
\usepackage{mathrsfs}

\usepackage[capitalize,noabbrev]{cleveref}

\theoremstyle{plain}

\theoremstyle{definition}

\theoremstyle{remark}

\usepackage[textsize=tiny]{todonotes}

\newcommand{\Mat}{\boldsymbol}

\newcommand{\real}{\mathbb{R}}

\DeclareMathOperator{\mean}{\mathbb{E}}
\DeclareMathOperator{\Var}{\mathbb{V}}
\DeclareMathOperator{\KL}{\mathcal{D}_{KL}}
\DeclareMathOperator{\gauss}{\mathcal{N}}
\DeclareMathOperator{\uniform}{\mathcal{U}}

\icmltitlerunning{SteinDreamer: Variance Reduction for Text-to-3D Score Distillation via Stein Identity}

\begin{document}

\twocolumn[
\icmltitle{SteinDreamer: Variance Reduction for Text-to-3D Score Distillation \\ via Stein Identity}

\icmlsetsymbol{equal}{*}
\icmlsetsymbol{intern}{*}

\begin{icmlauthorlist}
\icmlauthor{Peihao Wang}{ut,intern}
\icmlauthor{Zhiwen Fan}{ut}
\icmlauthor{Dejia Xu}{ut}
\icmlauthor{Dilin Wang}{meta}
\icmlauthor{Sreyas Mohan}{meta}
\icmlauthor{Forrest Iandola}{meta}
\icmlauthor{Rakesh Ranjan}{meta}
\icmlauthor{Yilei Li}{meta}
\icmlauthor{Qiang Liu}{ut}
\icmlauthor{Zhangyang Wang}{ut}
\icmlauthor{Vikas Chandra}{meta}
\end{icmlauthorlist}

\icmlaffiliation{ut}{University of Texas at Austin}
\icmlaffiliation{meta}{Meta Reality Labs}

\icmlkeywords{Text-to-3D Generative Models, Score Distillation, Variance Reduction, Stein's Method}

\vskip 0.1in

\centering\href{https://vita-group.github.io/SteinDreamer/}{\tt\small vita-group.github.io/SteinDreamer/}

\vskip 0.3in
]

\printAffiliationsAndNotice{\icmlIntershipWork}  %

\begin{abstract}
Score distillation has emerged as one of the most prevalent approaches for text-to-3D asset synthesis.
Essentially, score distillation updates 3D parameters by lifting and back-propagating scores averaged over different views.
In this paper, we reveal that the gradient estimation in score distillation is inherent to high variance.
Through the lens of variance reduction, the effectiveness of SDS and VSD can be interpreted as applications of various control variates to the Monte Carlo estimator of the distilled score.
Motivated by this rethinking and based on Stein's identity, we propose a more general solution to reduce variance for score distillation, termed \textit{Stein Score Distillation (SSD)}. SSD incorporates control variates constructed by Stein identity, 
allowing for arbitrary baseline functions. This enables us to include flexible guidance priors and network architectures to explicitly optimize for variance reduction.
In our experiments, the overall pipeline, dubbed \textit{SteinDreamer}, is implemented by instantiating the control variate with a monocular depth estimator.
The results suggest that SSD can effectively reduce the distillation variance and consistently improve visual quality for both object- and scene-level generation.
Moreover, we demonstrate that SteinDreamer achieves faster convergence than existing methods due to more stable gradient updates.
\end{abstract}

\section{Introduction}
\begin{figure}[t]
    \centering
    \includegraphics[width=0.9\linewidth]{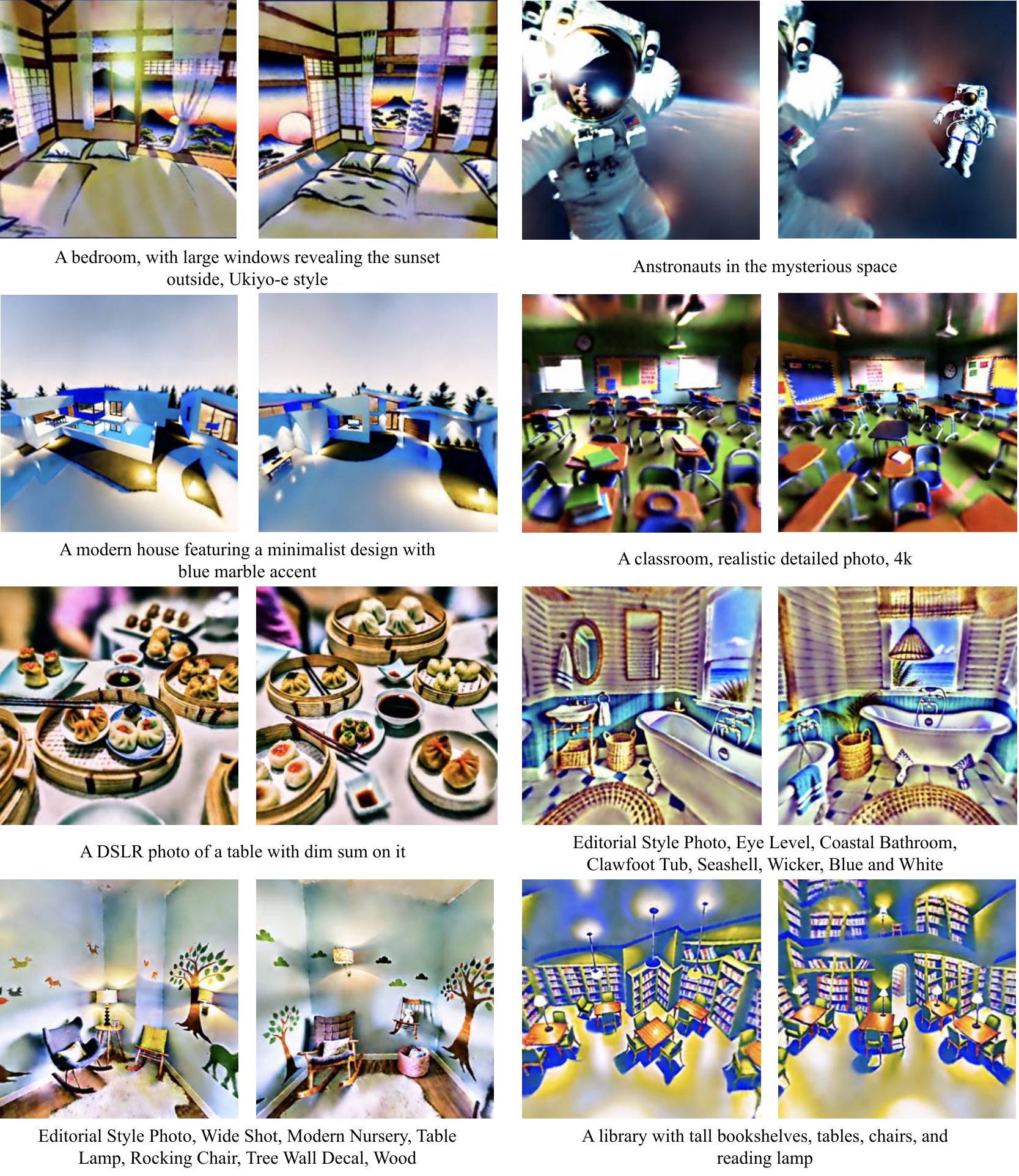}
    \vspace{-1em}
    \caption{\textbf{A gallery of text-to-3D results generated by SteinDreamer.} Our method can synthesize large-scale scenes with smooth geometries and rich textures according to complex text prompts. Zoom in for the best view.}
    \label{fig:var_vsd_sds}
    \vspace{-2em}
\end{figure}

There have been recent significant advancements in text-to-image generation, driven by diffusion models. Notable examples include \citet{nichol2021glide, ramesh2021zero, ramesh2022hierarchical} and \citet{sohl2015deep, ho2020denoising, song2019generative, song2020score, dhariwal2021diffusion}.
These developments have sparked growing interest in the realm of text-guided 3D generation. This emerging field aims to automate and accelerate 3D asset creation in the applications of virtual reality, movies, and gaming.
However, 3D synthesis poses significantly greater challenges.
Directly training generative models using 3D data, as explored in works by \citep{wu2016learning, yang2019pointflow, cai2020learning, nichol2022point, jun2023shap, chan2022efficient, shue20223d}, faces practical hurdles due to the scarcity of high-quality and diverse data. 
Moreover, the inherent complexity of generative modeling with 3D representations adds an extra layer of intricacy to this endeavor.

In recent times, techniques based on score distillation~\citep{poole2022dreamfusion, wang2023prolificdreamer}, exemplified by DreamFusion and ProlificDreamer, have gained prominence. 
These methods have garnered attention for their ability to effectively bypass the need for 3D data by leveraging a 2D diffusion model for 3D generation.
In particular, \citet{poole2022dreamfusion} introduces Score Distillation Sampling (SDS), which optimizes a differentiable 3D representation, such as NeRF \citep{mildenhall2020nerf}, by lifting and back-propagating image scores from a pre-trained text-to-image diffusion model.
Among its subsequent works \citep{lin2023magic3d, wang2023score, chen2023fantasia3d, metzer2023latent, wang2023taming}, ProlificDreamer stands out for significantly enhancing the generation quality through derived Variational Score Distillation (VSD)~\citep{wang2023prolificdreamer}.
In particular, VSD introduces an additional score for rendered image distribution that enhances the quality of distillation results.

Despite all these progresses, it is widely recognized that gradient obtained through score distillation techniques tend to be noisy and unstable due to the high uncertainty in the denoising process and the small batch size limited by computational constraints. Consequently, this leads to slow convergence and suboptimal solutions.
In this paper, we address this issue by proposing a unified variance reduction approach.
We reveal that both the noise term in SDS and the extra score function introduced by VSD have zero means, and thus can be regarded as \textit{control variates}.
The update of VSD is equivalent to the update of SSD in expectation. However, the gradient variance is smaller in VSD due to the effect of a better 
implementation the control variate.

Building on these insights, we present a more flexible control variate for score distillation, leveraging Stein identity \citep{stein1972bound, chen1975poisson, gorham2015measuring}, dubbed \textit{Stein Score Distillation (SSD)}.
Stein's identity, given by $\mean_{\Mat{x}\sim p}[\nabla\log p(\Mat{x})\cdot f(\Mat{x})^\top + \nabla_{\Mat{x}} f(\Mat{x})]=0$ for any distribution $p$ and function $f$ satisfying mild regularity conditions~\citep{stein1972bound, gorham2015measuring, liu2016kernelized}. This formulation establishes a broader class of control variates due to its zero means, providing flexibility in optimizing function $f$ for variance reduction.
Specifically, 
our \textit{Stein Score Distillation (SSD)} frames the distillation update as a combination of the score estimation from a pre-trained diffusion model and a control variate derived from Stein's identity. 
The first term aligns with that in SDS and VSD, serving to maximize the likelihood of the rendered image. 
The second control variate is tailored to specifically reduce gradient variance. 
Importantly, our construction allows us to incorporate arbitrary prior knowledge and network architectures in $f$, facilitating the design of control variates highly correlated with the lifted image score, leading to a significant reduction in gradient variance.

We integrate our proposed SSD into a text-to-3D generation pipeline, coined as \textit{SteinDreamer}.
Through extensive experiments, we demonstrate that SteinDreamer can consistently mitigate variance issues within the score distillation process.
For both 3D object and scene-level generation, SteinDreamer outperforms DreamFusion and ProlificDreamer by providing detailed textures, precise geometries, and effective alleviation of the Janus \citep{hong2023debiasing} and ghostly \citep{warburg2023nerfbusters} artifacts.
Lastly, it's worth noting that SteinDreamer, with its reduced variance, accelerates the convergence of 3D generation, reducing the number of iterations required by 14\%-22\%.

\section{Preliminaries}

\subsection{Score Distillation}

Diffusion models, as demonstrated by various works \citep{sohl2015deep, ho2020denoising, song2019generative, song2020score}, have proven to be highly effective in text-to-image generation.
Diffusion models learn a series of score functions $\nabla\log p_t(\Mat{x}_t | \Mat{y})$ for Gaussian perturbed image distribution $p_t(\Mat{x}_t | \Mat{y}) = \int \gauss(\Mat{x}_t | \alpha_t \Mat{x}_0, \sigma_t^2 \Mat{I}) p_0(\Mat{x}_0 | \Mat{y}) d\Mat{x}_0$, where $\alpha_t, \sigma_t > 0$ are annealing noise coefficients, and $\Mat{y}$ is the text embeddings.

Build upon the success of 2D diffusion models, \citet{poole2022dreamfusion, wang2023score, lin2023magic3d, chen2023fantasia3d, tsalicoglou2023textmesh, metzer2023latent, wang2023prolificdreamer, huang2023dreamtime} demonstrate the feasibility of using a 2D generative model to create 3D asserts.
Among these works, score distillation techniques play a central role by providing a way to guide a differentiable 3D representation using a pre-trained text-to-image diffusion model. 
Essentially, score distillation lifts and back-propagates signals estimated from a 2D prior to update a differentiable 3D representation, such as Neural Radiance Field (NeRF) \citep{mildenhall2020nerf}, via the chain rule \citep{wang2023score}.
There are primarily two types of distillation schemes elaborated below:

\paragraph{Score Distillation Sampling.}
The main idea of \textit{Score Distillation Sampling (SDS)} is to utilize a score function of some pre-trained 2D image distribution $\nabla \log p_t$ and the denoising score matching loss to optimize a 3D representation, denoted as $\Mat{\theta}$, such that it semantically matches a given text prompt based on its multi-view projections.
By taking derivatives with respect to 3D parameters $\Mat{\theta}$ and dropping the Jacobian matrix of the score function, SDS yields the following update rule\footnote{By default, Jacobian matrices are transposed.}:
\begin{align} \label{eqn:sds}
\Mat{\Delta}_{SDS} = \mean\left[ \omega(t) \frac{\partial g(\Mat{\theta}, \Mat{c})}{\partial \Mat{\theta}} \left(\sigma_t \nabla \log p_t(\Mat{x}_t | \Mat{y}) - \Mat{\epsilon}\right) \right],
\end{align}
where the expectation is taken over time step $t \sim \uniform[0, T]$, camera pose $\Mat{c} \sim p_c$, and white noise $\Mat{\epsilon} \sim \gauss(\Mat{0}, \Mat{I})$. The noisy input of score function is denoted as $\Mat{x}_t = \alpha_t g(\Mat{\theta}, \Mat{c}) + \sigma_t \Mat{\epsilon}$ and $\omega(t) > 0$ are time-dependent coefficients. $g(\Mat{\theta}, \Mat{c})$ renders a 2D view from $\Mat{\theta}$ given $\Mat{c}$.
In this work, we represent $\Mat{\theta}$ as a NeRF, wherein $g(\Mat{\theta}, \Mat{c})$ represents a volumetric renderer displaying each image pixel under camera pose $\Mat{c}$ by performing ray tracing based rendering \citep{max1995optical}.
Meanwhile, $-\sigma_t \nabla \log p_t$ can be surrogated by a noise estimator from a pre-trained diffusion model.

\paragraph{Variational Score Distillation.}
ProlificDreamer introduced a new variant of score distillation, \textit{Variational Score Distillation (VSD)}~\citep{wang2023prolificdreamer}, through the lens of particle-based variational inference \citep{liu2016stein, liu2017stein, detommaso2018stein}.
ProlificDreamer minimizes the KL divergence between $p_t(\Mat{x})$ and the image distribution rendered from a 3D representation $\Mat{\theta}$. 
The authors derive the following update rule through Wasserstein gradient flow:
\begin{align} \label{eqn:vsd}
\Mat{\Delta}_{VSD} = \mean \left[ \omega(t) \frac{\partial g(\Mat{\theta}, \Mat{c})}{\partial \Mat{\theta}} (\sigma_t \nabla\log p_t(\Mat{x}_t | \Mat{y}) \right. \nonumber \\
\left. \vphantom{\frac{\partial g(\Mat{\theta}, \Mat{c})}{\partial \Mat{\theta}}} - \sigma_t \nabla\log q_t(\Mat{x}_t | \Mat{c})) \right],
\end{align}
where the expectation is taken over all relevant variables by default.
Notably, there emerges a new score function of probability density function $q_t(\Mat{x} | \Mat{c})$,
which characterizes the conditional distribution of noisy rendered images given the camera pose $\Mat{c}$.
While $\nabla\log p_t$ can be approximated in a similar manner using an off-the-shelf diffusion model, $\nabla\log q_t$ is not readily available.
The solution provided by \citet{wang2023prolificdreamer} is to fine-tune a pre-trained diffusion model using the rendered images. The approach results an alternating optimization paradigm between objective Eq. \ref{eqn:vsd} and additional score matching loss:
\begin{align}
\min_{\nabla \log q_t} \mean_{t, \Mat{c}, \Mat{\epsilon} \sim \gauss(\Mat{0}, \Mat{I})} \left[ \omega(t) \lVert \sigma_t \nabla \log q_t(\Mat{x}_t | \Mat{c}) - \Mat{\epsilon} \rVert_2^2 \right], \nonumber
\end{align}
where $\nabla \log q_t(\Mat{x}_t | \Mat{c})$ is a diffusion model initialized with the pre-trained $\nabla\log p_t(\Mat{x}_t | \Mat{y})$, parameterized by LoRA \citep{hu2021lora}, and additionally conditioned on camera pose.

\begin{figure}[t]
    \centering
    \includegraphics[width=0.9\linewidth]{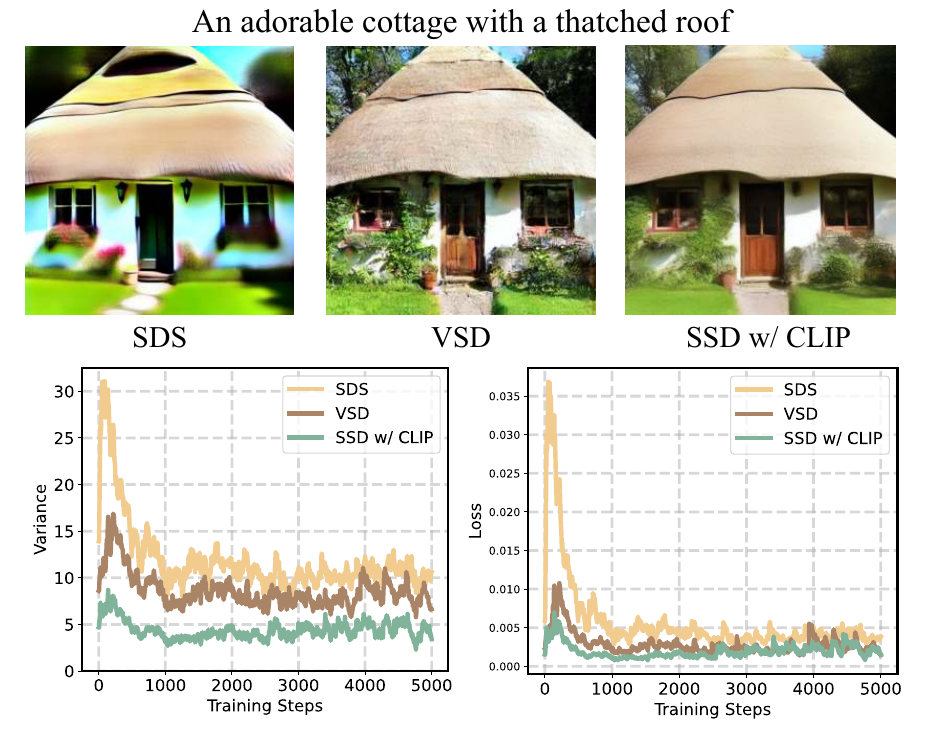}
    \vspace{-1em}
    \caption{\textbf{Comparison between SDS, VSD, and SSD on 2D space.}  We monitor the variance of $\Mat{\Delta}_{SDS}$ $\Mat{\Delta}_{VSD}$, and $\Mat{\Delta}_{SSD}$ for every 100 training step. We show that variance level is highly correlated to the final performance and convergence speed.}
    \label{fig:var_vsd_sds}
    \vspace{-1em}
\end{figure}

\subsection{Control Variate} \label{sec:control_var}

Later in this work, we will introduce control variate into the context of score distillation. 
Control variate is a widely utilized technique to reduce variance for Monte Carlo estimator in various fields, including physical simulation
\citep{davies2004high}, graphical rendering \citep{kajiya1986rendering, muller2020neural}, network science \citep{meyn2008control, chen2017stochastic}, and reinforcement learning \citep{williams1992simple, sutton1998introduction, sutton1999policy, liu2017action}.
Suppose we want to estimate the expectation $\mean_{\Mat{x} \sim q(\Mat{x})}[f(\Mat{x})]$ for some function $f$ via Monte Carlo samples $\{ \Mat{x}_i \in \real^D \}_{i=1}^{N}$: $\Mat{\Delta} = \frac{1}{N} \sum_{i=1}^{N} f(\Mat{x}_i)$.
The estimator $\Mat{\Delta}$ is supposed to have large variance when $N$ is small.
Consider we have control variate as a function $h$ with analytic mean under $q(\Mat{x})$, which can be assumed to equal zero without loss of generality.
Then we can construct an unbiased estimator by adding term $\Mat{\xi} = \frac{1}{N} \sum_{i=1}^{N} h(\Mat{x}_i)$: $\Mat{\Delta}^{\dag} = \frac{1}{N} \sum_{i=1}^{N} (f(\Mat{x}_i) + \Mat{\mu} \odot h(\Mat{x}_i))$, where $\Mat{u} \in \real^D$ is a group of reweighting coefficients and $\odot$ denotes element-wise multiplication.
The resultant estimator has variance for the $i$-th entry:
\begin{align}
\Var[ \Mat{\Delta}^{\dag}_i ] = \Var\left[ \Mat{\Delta}_i \right] + \Mat{\mu}_i^2 \Var\left[ \Mat{\xi}_i \right] + 2 \Mat{\mu}_i \mean[\Mat{\Delta} \Mat{\xi}^\top]_{ii}.
\end{align}
It is possible to reduce $\Var[ \Mat{\Delta}^{\dag}_i]$ by selecting $h$ and $\Mat{u}$ properly.
To maximize variance reduction, $\Mat{u}$ is chosen as $-\mean[\Mat{\Delta} \Mat{\xi}^\top]_{ii} / \Var[ \Mat{\xi}_i ]$, leading to $\Var[ \Mat{\Delta}^{\dag}_i] = (1 - \operatorname{Corr}(\Mat{\Delta}_i, \Mat{\xi}_i)^2) \Var[ \Mat{\Delta}_i]$, where $\operatorname{Corr}(\cdot, \cdot)$ denotes the correlation coefficient.
This signifies that higher correlation between functions $f$ and $h$ leads to higher variance reduction.

\section{Rethinking SDS and VSD: A Control Variate Perspective} \label{sec:rethink}

In this section, we reveal that the variance of update estimation may play a key role in score distillation.
At first glance, SDS and VSD differ in their formulation and implementation.
However, our first theoretical finding reveals that SDS and (single-particle) VSD are equivalent in their expectation, i.e., 
$\Mat{\Delta}_{SDS} = \Mat{\Delta}_{VSD}$.
We formally illustrate this observation below.

As a warm-up, we inspect SDS via the following rewriting.
\begin{align} \label{eqn:decomp_sds_f}
\Mat{\Delta}_{SDS} = &\mean_{t, \Mat{c}, \Mat{\epsilon}} \underbrace{\left[ \omega(t) \frac{\partial g(\Mat{\theta}, \Mat{c})}{\partial \Mat{\theta}} \sigma_t \nabla \log p_t(\Mat{x}_t | \Mat{y}) \right]}_{f(t, \Mat{\theta}, \Mat{x}_t, \Mat{c})} \\
\label{eqn:decomp_sds_h} & - \mean_{t, \Mat{c}, \Mat{\epsilon}} \underbrace{\left[ \omega(t) \frac{\partial g(\Mat{\theta}, \Mat{c})}{\partial \Mat{\theta}} \Mat{\epsilon} \right]}_{h_{SDS}(t, \Mat{\theta}, \Mat{x}, \Mat{c})}.
\end{align}
where the second term $\mean[h_{SDS}(t, \Mat{\theta}, \Mat{x}, \Mat{c})] = \Mat{0}$ simply because it is the expectation of a zero-mean Gaussian vector.
For VSD, we follow a similar derivation and obtain:
\begin{align} \label{eqn:decomp_vsd_f}
\Mat{\Delta}_{VSD} = & \mean_{t, \Mat{c}, \Mat{\epsilon}} \underbrace{\left[ \omega(t) \frac{\partial g(\Mat{\theta}, \Mat{c})}{\partial \Mat{\theta}} \sigma_t \nabla \log p_t(\Mat{x}_t | \Mat{y}) \right]}_{f(t, \Mat{\theta}, \Mat{x}, \Mat{c})} \\
\label{eqn:decomp_vsd_h} & - \mean_{t, \Mat{c}, \Mat{\epsilon}} \underbrace{\left[ \omega(t) \frac{\partial g(\Mat{\theta}, \Mat{c})}{\partial \Mat{\theta}} \sigma_t \nabla \log q_t(\Mat{x}_t | \Mat{c}) \right]}_{h_{VSD}(t, \Mat{\theta}, \Mat{x}, \Mat{c})}.
\end{align}
Taking a closer look, we find that the second term is also zero: $\mean[h_{VSD}(t, \Mat{\theta}, \Mat{x}, \Mat{c})] = \Mat{0}$.
This can be proven by showing that $q_t(\Mat{x} | \Mat{c})$ turns out to be a zero-mean Gaussian distribution or applying the inverse chain rule followed by the fact that the first-order moment of the score function is constantly zero.
Moreover, the first term $\mean[f(t, \Mat{\theta}, \Mat{x}, \Mat{c})]$ of both SDS and VSD equals to $-\nabla_{\Mat{\theta}} \mean_t \left[\KL(q_{t}(\Mat{x}_t | \Mat{c}) \Vert p_t(\Mat{x}_t | \Mat{y}))\right]$. This implies that SDS and VSD are equivalent to gradient descent algorithms, minimizing the distribution discrepancy between the noisy rendered image distribution and the Gaussian perturbed true image distribution.
We refer interested readers to Appendix \ref{sec:app_proofs} for  the full derivation.

However, in most scenarios, 
empirical evidence indicates that VSD consistently outperforms SDS, despite both methods aiming to minimize the same objective. 
To explain this paradox, we posit that the underlying source of their performance disparities is attributed to the \textit{variance} of stochastic simulation of the expected updates by SDS and VSD.
The numerical evaluation of Eq. \ref{eqn:sds} and Eq. \ref{eqn:vsd} typically 
relies on Monte Carlo estimation over a mini-batch.
Unfortunately, rendering a full view from NeRF and performing inference with diffusion models are computationally demanding processes, leading to a constrainted number of rendered views  within a single optimization step - often limited to just one, as suggested in previous work \citep{poole2022dreamfusion}.
Additionally, the term related to the score function within the expectation undergoes a denoising procedure, notorious for its instability and high uncertainty, especially when $t$ is large.
Hence, despite SDS and VSD having identical means, we argue that the variance of their numerical estimation significantly differs.

We empirically validate this speculation in Fig. \ref{fig:var_vsd_sds}, wherein we utilize SDS add VSD to sample 2D images from a pre-trained diffusion model, akin to \citet{wang2023prolificdreamer}.
We visualize the variance of $\Mat{\Delta}_{SDS}$ and $\Mat{\Delta}_{VSD}$ during the training process. 
It can be observed that VSD converges faster and yields results with richer details.
Moreover, Fig. \ref{fig:var_vsd_sds} demonstrates a clear separation between SDS and VSD in terms of the stochastic gradients' variance.
Such phenomenon signifies the variance of gradient estimation is highly correlated to the resultant performance.

To gain insight into the variance disparity between SDS and VSD, 
we connect SDS and VSD via the concept of control variates.
As introduced in Sec. \ref{sec:control_var}, a control variate is a zero-mean random variable capable of reducing the variance of Monte Carlo estimator when incorporated into the simulated examples.
Notably, both $h_{SDS}(t, \Mat{\theta}, \Mat{x}, \Mat{c})$ and $h_{VSD}(t, \Mat{\theta}, \Mat{x}, \Mat{c})$ can be regarded as control variates,  as confirmed by Eq. \ref{eqn:decomp_sds_h} and Eq. \ref{eqn:decomp_vsd_h} due to their zero means.
Consequently, SDS and VSD can be interpreted as Monte Carlo estimators of the gradient of the KL divergence, integrated with different control variates.
As demonstrated in Sec. \ref{sec:control_var}, control variate with higher correlation to the estimated variable leads to larger variance reduction.
VSD exhibits lower variance primarily because $\nabla\log q_t(\Mat{x} | \Mat{c})$ in control variate $h_{VSD}$ is fine-tuned from $\nabla\log p_t(\Mat{x} | \Mat{c})$, and perhaps resulting in higher correlation compared to the pure Gaussian noises in $h_{SDS}$.

\begin{figure*}[t]
    \centering
    \includegraphics[width=\textwidth]{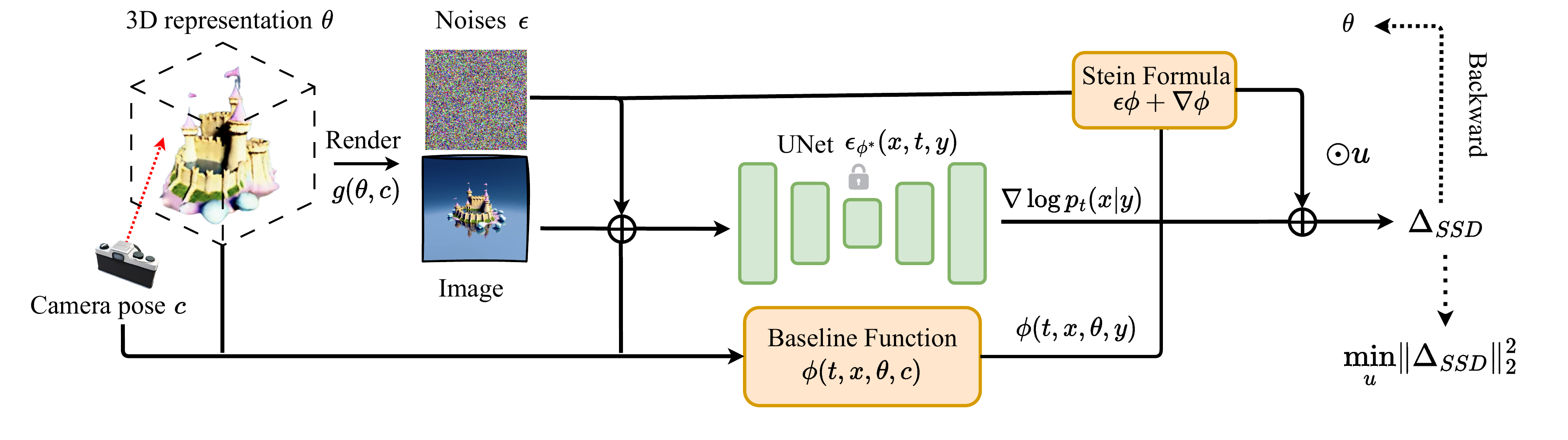}
    \vspace{-2mm}
    \caption{\small \textbf{Pipeline of SteinDreamer.} We incorporate control variates constructed by Stein's identity into a score distillation pipeline, allowing for arbitrary baseline functions. In practice, we implement the baseline functions with a monocular depth or normal estimator.}
    \label{fig:pipeline}
\end{figure*}

\section{Stein Score Distillation}

\looseness=-1
Having revealed that variance control is one of the key knobs to improve the performance of score distillation, we extend the family of control variates that can be used for score distillation in this section.

\subsection{Stein Control Variates for Score Distillation} \label{sec:stein_cv}

Our main inspiration is drawn from \citet{oates2017control, liu2017stein, roeder2017sticking} that Stein's identity can be served as a powerful and flexible tool to construct zero-mean random variables.
We consider Stein's identity associated with any conditional probability $p(\Mat{x}_t | \Mat{\theta}, \Mat{c})$ as below:
\begin{align} 
\mean_{\Mat{x}_t \sim p(\Mat{x}_t | \Mat{\theta}, \Mat{c})} \left[ \nabla \log p(\Mat{x}_t | \Mat{\theta}, \Mat{c}) \phi(t, \Mat{\theta}, \Mat{x}_t, \Mat{c}) \right. \nonumber \\ 
\label{eqn:stein_id} \left. + \nabla_{\Mat{x}_t} \phi(t, \Mat{\theta}, \Mat{x}_t, \Mat{c}) \right] = \Mat{0},
\end{align}
where $\phi(t, \Mat{\theta}, \Mat{x}_t, \Mat{c})$ is referred to as the \textit{baseline function}, which can be arbitrary scalar-value function satisfying regularity conditions \citep{stein1972bound, gorham2015measuring, liu2016kernelized}.
By plugging $q_t(\Mat{x}_t | \Mat{\theta}, \Mat{c})$ into Eq. \ref{eqn:stein_id}, we can construct our control variate $h_{SSD}(t, \Mat{\theta}, \Mat{c}, \Mat{x}_t)$ as follows:
\begin{align} \label{eqn:ssd_cv}
h_{SSD} = \omega(t) \frac{\partial g(\Mat{\theta}, \Mat{c})}{\partial \Mat{\theta}} \bigg[\Mat{\epsilon} \phi(t, \Mat{\theta}, \Mat{x}_t, \Mat{c})
+ \nabla_{\Mat{x}_t} \phi(t, \Mat{\theta}, \Mat{x}_t, \Mat{c}) \bigg],
\end{align}
where $\Mat{x}_t = \alpha_t g(\Mat{\theta}, \Mat{c}) + \sigma_t \Mat{\epsilon}$ and $\Mat{\epsilon} \sim \gauss(\Mat{0}, \Mat{I})$.
Additional details and derivations are provided in Appendix \ref{sec:app_proofs}.
The advantage of $h_{SSD}$ lies in its flexibility to define an infinite class of control variates, characterized by arbitrary baseline function $\phi(t, \Mat{\theta}, \Mat{x}_t, \Mat{c})$.

\subsection{Variance Minimization via Stein Score Distillation} \label{sec:stein_cv}

We propose to adopt $h_{SSD}$ as the control variate for score distillation.
In addition to $h_{SSD}$, we introduce a group of learnable weights $\Mat{\mu} \in \real^{D}$ to facilitate optimal variance reduction following the standard scheme introduced in Sec. \ref{sec:control_var}.
Altogether, we present the following update rule, termed as \textit{Stein Score Distillation (SSD)}:
\begin{align}
\Mat{\Delta}_{SSD} = \mean_{t, \Mat{c}, \Mat{\epsilon}} \left[ \omega(t) \frac{\partial g(\Mat{\theta}, \Mat{c})}{\partial \Mat{\theta}} (\sigma_t \nabla \log p_t(\Mat{x}_t | \Mat{y}) \right. \nonumber \\
\label{eqn:ssd} + \left.\vphantom{\frac{\partial g(\Mat{\theta}, \Mat{c})}{\partial \Mat{\theta}}} \Mat{\mu} \odot [\Mat{\epsilon} \phi(t, \Mat{\theta}, \Mat{x}_t, \Mat{c}) + \nabla_{\Mat{x}_t} \phi(t, \Mat{\theta}, \Mat{x}_t, \Mat{c})] ) \right].
\end{align}
Here $\phi(t, \Mat{\theta}, \Mat{x}_t, \Mat{c})$ can be instantiated using any neural network architecture taking 3D parameters, noisy rendered image, and camera pose as the input.

In our experiments, we employ a pre-trained monocular depth estimator, MiDAS \citep{ranftl2020towards, ranftl2021vision}, coupled with domain-specific loss functions to construct $\phi(t, \Mat{\theta}, \Mat{x}_t, \Mat{c})$, as a handy yet effective choice. Specifically: 
\begin{align} \label{eqn:midas_phi}
\phi(t, \Mat{x}_t, \Mat{\theta}, \Mat{c}) = -\ell (\alpha(\Mat{\theta}, \Mat{c}), \operatorname{MiDAS}(\Mat{x}_t)).
\end{align}
Here $\operatorname{MiDAS}(\cdot)$ can estimate either depth or normal map from noisy observation $\Mat{x}_t$. And $\alpha(\cdot, \cdot)$ is chosen as the corresponding depth or normal renderer of the 3D representation $\Mat{\theta}$, and $\ell(\cdot, 
\cdot)$ is the Pearson correlation loss when estimating depth map 
or cosine similarity loss when considering normal map.

As introduced in Sec. \ref{sec:control_var}, there exists a closed-form $\Mat{\mu}$ that maximizes the variance reduction.
However, it assumes the correlation between the control variate and the random variable of interest is known.
Instead, we propose to directly optimize variance by adjusting $\Mat{\mu}$ to minimize the second-order moment of Eq. \ref{eqn:ssd} since its first-order moment is independent of 
$\Mat{\mu}$:
\begin{align} \label{eqn:var_min}
\min_{\Mat{\mu}} &\mean_{t, \Mat{c}, \Mat{\epsilon}} \left[\left\lVert \omega(t) \frac{\partial g(\Mat{\theta}, \Mat{c})}{\partial \Mat{\theta}} (\sigma_t \nabla \log p_t(\Mat{x}_t | \Mat{y}) \right.\right. \\  
& \left.\vphantom{[} \left.\vphantom{\frac{\partial g(\Mat{\theta}, \Mat{c})}{\partial \Mat{\theta}}} + \Mat{\mu} \odot [\Mat{\epsilon} \phi(t, \Mat{\theta}, \Mat{x}_t, \Mat{c}) + \nabla_{\Mat{x}_t} \phi(t, \Mat{\theta}, \Mat{x}_t, \Mat{c})]) \right\rVert_2^2\right], \nonumber
\end{align}
which essentially imposes a penalty on the gradient norm of $\Mat{\theta}$.
We alternate between optimizing $\Mat{\theta}$ and $\Mat{\mu}$ using SSD gradient in Eq. \ref{eqn:ssd} and the objective function in Eq. \ref{eqn:var_min}, respectively.
We refer to our complete text-to-3D framework as \textit{SteinDreamer}, and its optimization paradigm is illustrated in Fig. \ref{fig:pipeline}.

Specifically, during each optimization iteration, SteinDreamer performs the following steps:
1) renders RGB map and depth/normal map from a random view of $\Mat{\theta}$, 2) perturbs the RGB map and obtains the score estimation using a pre-trained diffusion model and monocular depth/normal prediction from a pre-trained MiDAS, 3) computes $\phi$ via Eq. \ref{eqn:midas_phi} and its gradient via auto-differentiation to form control variate $h_{SSD}$, 4) weights the control variate by $\Mat{\mu}$ and combine it with the diffusion score $\nabla \log p_t(\Mat{x}_t | \Mat{y})$, 5) back-propagates $\Mat{\Delta}_{SSD}$ through the chain rule to update 3D parameters $\Mat{\theta}$.
In the other fold, SteinDreamer keeps $\Mat{\theta}$ frozen and optimizes $\Mat{\mu}$ to minimize the $\ell_2$ norm of the update signals on $\Mat{\theta}$ according to Eq. \ref{eqn:var_min}.

\begin{figure*}[t]
    \centering
    \includegraphics[width=0.9\textwidth]{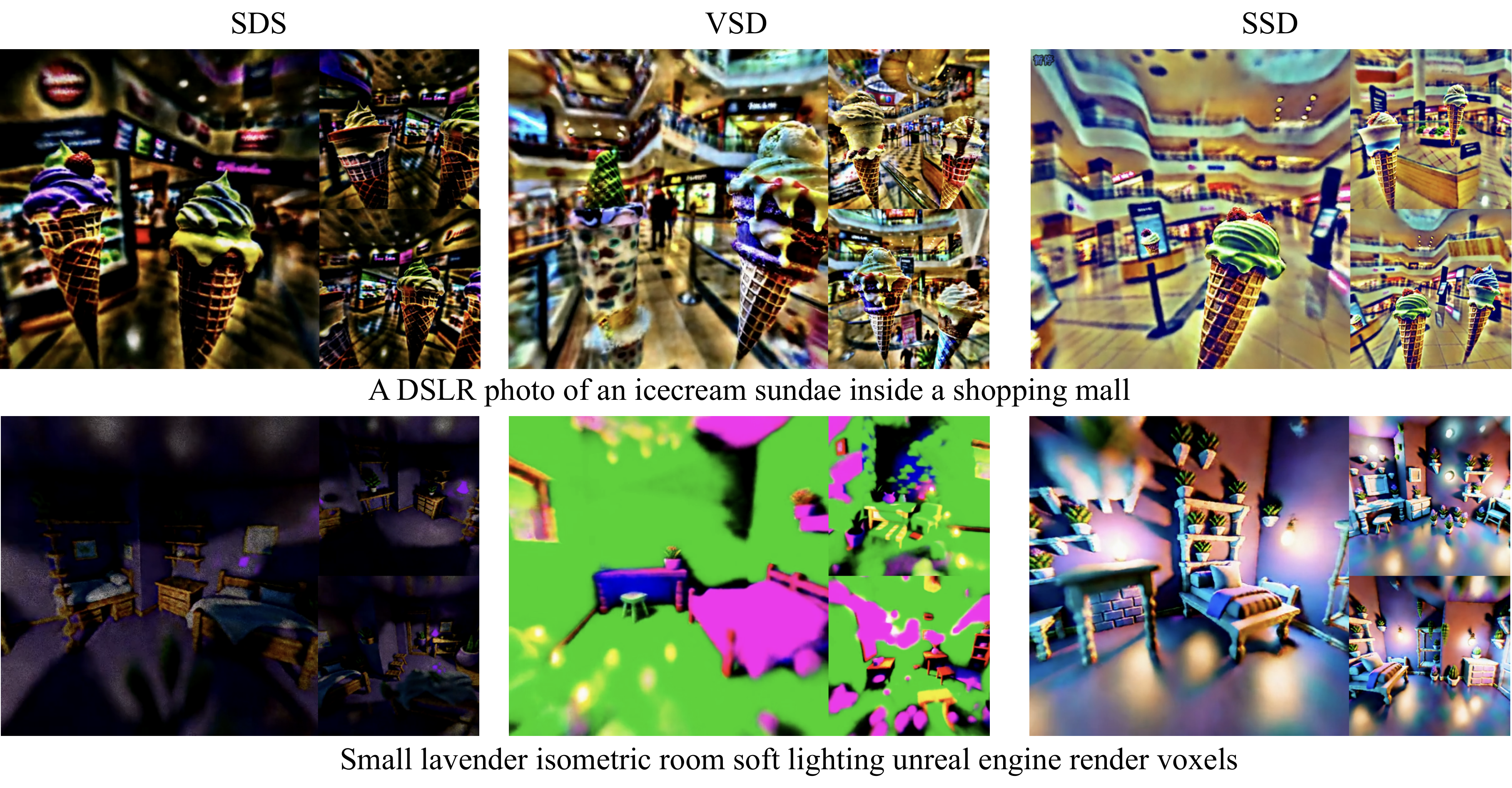}
    \vspace{-1em}
    \caption{\textbf{Scene-level qualitative comparisons.} Compared to existing methods, SteinDreamer w/ normal estimator presents more realistic textures with better details.}
    \label{fig:scene_res_steinnormal}
    \vspace{-1em}
\end{figure*}

\subsection{Discussion} \label{sec:discuss}

\paragraph{Comparison with SDS and VSD.}
First, SDS is a special case of SSD when taking $\phi(t, \Mat{\theta}, \Mat{x}_t, \Mat{c}) = -1$.  
This observation suggests the potential for SSD to provide a lower variance in gradient estimation due to its broader range in representing control variates.
As demonstrated in \citet{oates2017control}, 
an optimal control variate can be constructed using Stein's identity by carefully selecting $\phi$, achieving a zero-variance estimator.
The key advantage of SSD lies in its flexibility in choosing the baseline function $\phi$, which can directly condition and operate on all relevant variables.
Furthermore, the expressive power of SSD surpasses that of VSD, in which $\nabla \log q_t(\Mat{x}_t | \Mat{c})$ implicitly conditions on $\Mat{\theta}$ through $\Mat{x}_t$ and $\Mat{c}$.
To validate our argument, we present an initial experiment on the 2D space in Fig. \ref{fig:var_vsd_sds}, in which we construct a baseline function using CLIP loss \citep{radford2021learning} to regularize image sampling.
Please see more details in Appendix \ref{sec:app_expr_details}.
We show that SSD-based stochastic gradient incurs lower variance than the other two baselines, which results in eye-pleasing results and faster convergence.
We defer more technical comparison to Appendix \ref{sec:app_discuss}.

\paragraph{SVGD-Induced Score Distillation.}
\citet{kim2023collaborative} proposes Collaborative Score Distillation (CSD) to sample latent parameters via Stein Variational Gradient Descent (SVGD).
While both methods are grounded in Stein's method, the underlying principles significantly differ. In CSD, the SVGD-based update takes the form of the Stein discrepancy: $\max_{\phi \in \mathcal{F}}\mathbb{E}_{\Mat{x} \sim q(\Mat{x})} [\phi(\Mat{x}) \nabla \log p(\Mat{x}) + \nabla_{\Mat{x}} \phi(\Mat{x})]$, where $\phi(\Mat{x})$ is often interpreted as an update direction constrained by a function class $\mathcal{F}$ (RBF kernel space in \citet{kim2023collaborative}). In contrast, our update rule appends a zero-mean random variable via the Stein identity after the raw gradient of the KL divergence (Eq. \ref{eqn:ssd}), where $\phi(\Mat{x})$ typically represents a pre-defined baseline function. The potential rationale behind CSD to reducing variance lies in introducing the RBF kernel as a prior to constrain the solution space by modeling pairwise relations between data samples. Our SSD is centered around constructing a more general control variate that correlates with the random variable of interest, featuring zero mean but variance reduction.

\begin{figure*}[t]
    \centering
     \includegraphics[width=\textwidth]{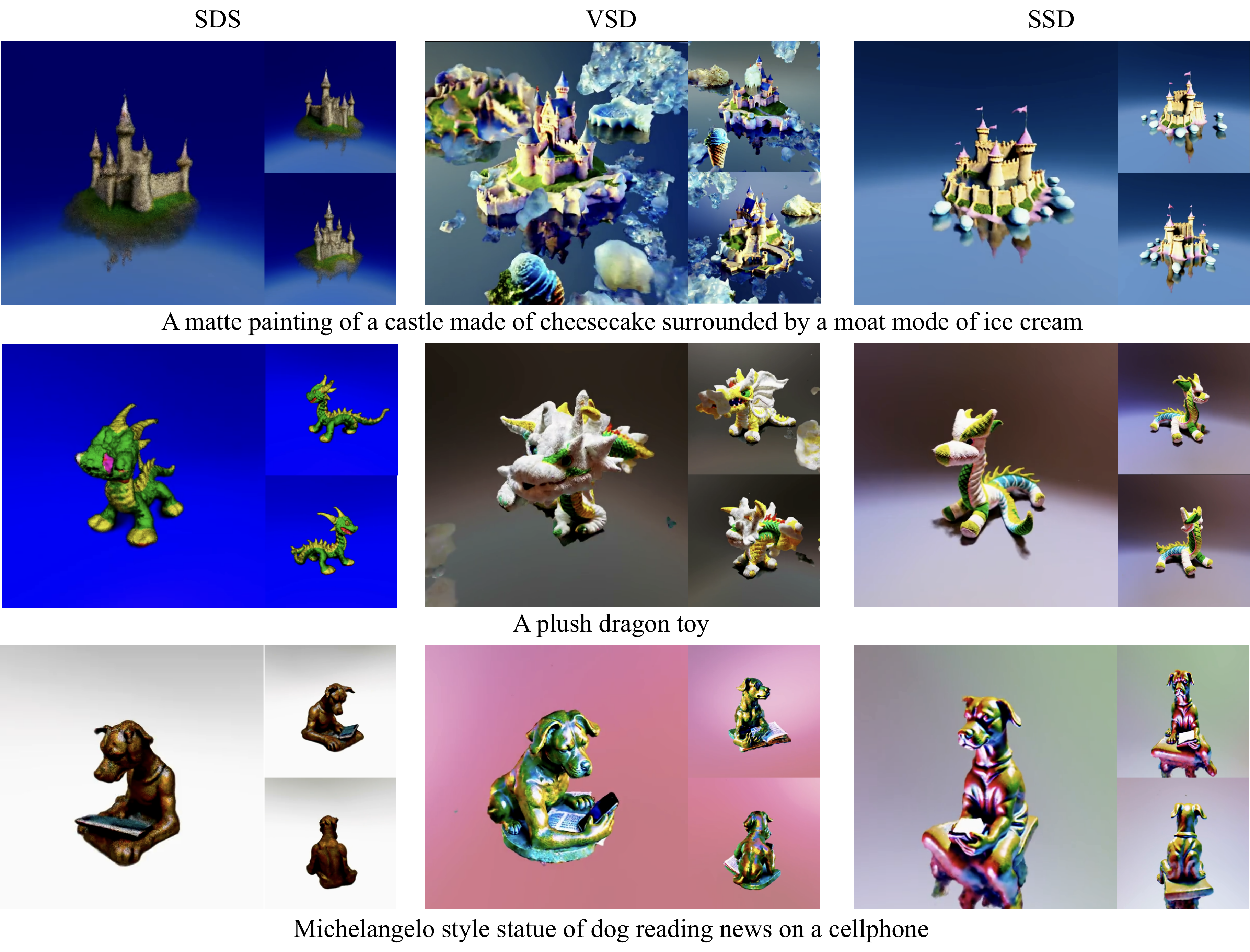}
    \vspace{-2em}
    \caption{\textbf{Object-level qualitative comparisons.} Compared to existing methods, our SteinDreamer w/ normal estimator delivers smoother geometry, more detailed texture, and fewer floater artifacts.}
    \label{fig:obj_res_steinnormal}
    \vspace{-1em}
\end{figure*}

\begin{figure*}[t]
    \centering
    \includegraphics[width=\textwidth]{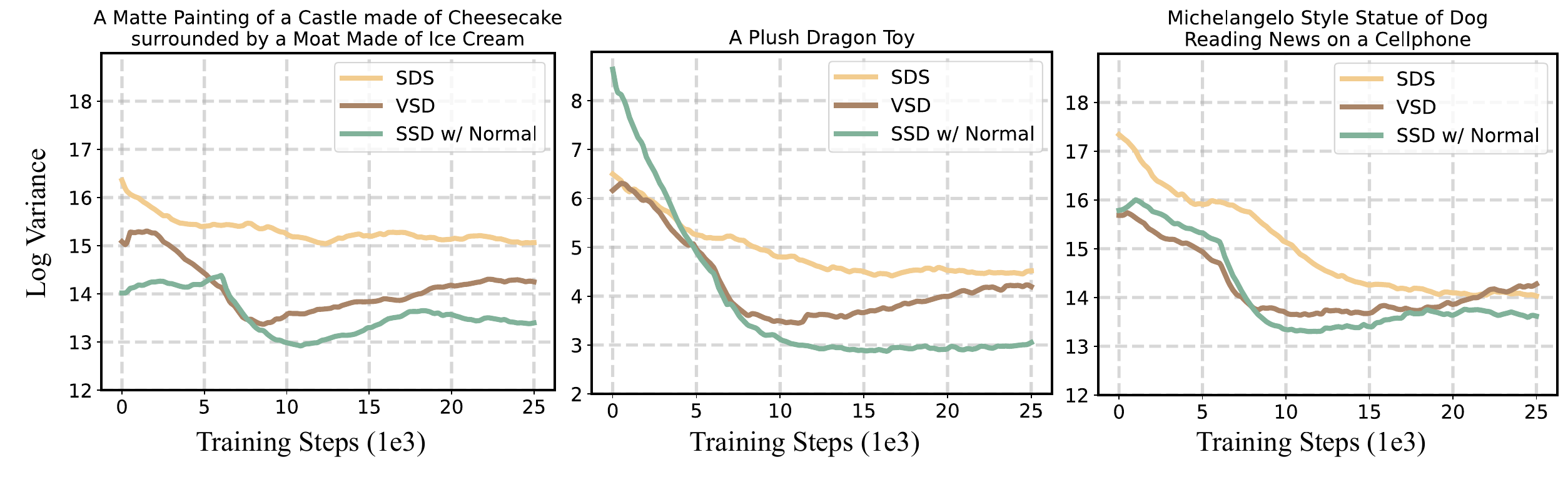}
    \vspace{-2em}
    \caption{\small \textbf{Variance comparison of $\Mat{\Delta}_{SDS}$, $\Mat{\Delta}_{VSD}$, and $\Mat{\Delta}_{SSD}$ with normal estimator.}  We visualize how the variance of the investigated three methods for every 1,000 steps. The variance decays as the training converges while $\Mat{\Delta}_{SSD}$ consistently achieves lower variance throughout the whole process. }
    \label{fig:var_vsd_sds_ssd_normal}
    \vspace{-2em}
\end{figure*}

\section{Experiments}

\label{sec:expr_res}

We conduct experiments for both scene-level and object-level text-to-3D generation.
The text prompts utilized in object generation are collected from \citet{wang2023prolificdreamer} while those for scene generation are originally from \citet{hollein2023text2room}, and \citet{zhang2023scenewiz3d}.
We mainly compare against the seminal works SDS from DreamFusion and VSD from ProlificDreamer.
For a fair comparison, we utilize the open-source  threestudio~\footnote{\url{https://github.com/threestudio-project/threestudio}} as a unified benchmarking implementation.
We thoroughly test our proposed SteinDreamer with both depth estimator and normal estimator priors.
In the main text, we mainly present results produced by SteinDreamer with normal estimator while deferring results of SteinDreamer with depth prior to Appendix \ref{sec:more_scene_res}.
All training hyper-parameters are kept the same with ProlificDreamer.
For simplicity, we evaluate VSD with the particle number equal to one.

\subsection{Qualitative Evaluation}

\paragraph{Large Scene Generation.}
We evaluate the performance of our method for large-scale scene generation.
The detailed comparisons with baselines on 360\textdegree~scene-level generation is presented in Fig.~\ref{fig:scene_res_steinnormal} and Fig.~\ref{fig:scene_res_steindepth}.
SDS delivers blurry results with unrealistic colors and textures.
The results from VSD suffer from the noisy background, and we also observe that the VSD loss can diverge in the texture refining process (Fig.~\ref{fig:scene_res_steinnormal} ).
In comparison, we observe that results generated by SteinDreamer are much sharper in appearance and enjoy better details.

\paragraph{Object Centric Generation.}
We exhibit our object-level qualitative results in Fig.~\ref{fig:obj_res_steinnormal}  and Fig.~\ref{fig:obj_res_steindepth} for SteinDreamer with normal or depth prior, respectively.
Compared with SDS, our SteinDreamer presents novel views with less over-saturation and over-smoothing artifacts.
When comparing with VSD, not only does our SteinDreamer generate smoother geometry, but also delivers sharper textures without contamination of floaters.
Additionally, it is worth noting that our SteinDreamer can potential alleviate the ``Janus'' problem by incorporating correct geometric priors, as shown in the dog statue case.
We further monitor the variance for all the demonstrated examples during the training stage in Fig.~\ref{fig:var_vsd_sds_ssd_normal}.
It is clear that our SteinDreamer consistently has lower variance than compared baselines throughout the course of training.
This observation supports our theory on the correlation between variance and synthesis quality.

\begin{table}[!t]
\centering
\resizebox{\linewidth}{!}{\begin{tabular}{lccccc}
\toprule
\multirow{2}{*}{Methods} & \multicolumn{2}{c}{Scene-level} &   \multicolumn{2}{c}{Object-level} \\
\cline{2-5}
& CLIP ($\downarrow$) & FID ($\downarrow$) & CLIP ($\downarrow$) & FID ($\downarrow$) \\
\hline
SDS \citep{poole2022dreamfusion} & 0.848\textpm0.068 & 298.49\textpm57.63 & 0.898\textpm0.133 & 282.50\textpm52.11 \\
VSD \citep{wang2023prolificdreamer} &   0.800\textpm0.051 & 268.92\textpm49.65 & 0.763\textpm0.100 & 271.62\textpm62.77 \\
SSD (Ours) & \textbf{0.762\textpm0.039} &      \textbf{240.17\textpm45.54} & \textbf{0.720\textpm0.064} & \textbf{251.31\textpm49.70 } \\
\bottomrule
\end{tabular}}
\caption{\textbf{Quatitative results.} We compare the CLIP ad FID distance ($\downarrow$ the lower the better) of demonstrated results among different approaches. Best results are marked in \textbf{bold} font.}
\label{tab:numerical}
\vspace{-1em}
\end{table}

\subsection{Quantitative Analysis}
\label{sec:quant_res}

Additionally, we report numerical comparison of our methods against other baselines in Tab. \ref{tab:numerical}.
We adopt CLIP distance \citep{xu2022neurallift} and FID \citep{heusel2017gans} as the metrics.
For scene generation, we consider 12 text prompts used throughout the whole paper.
For object-level generation, we borrow 20 text prompts from \citet{wang2023prolificdreamer}.
For each scene or object, we run each algorithm for three times.
The reported results is an average of all generated results.
We observe that our method consistently outperforms SDS and VSD with a significant gap.

\subsection{Ablation Studies}
\begin{figure}[!t]
 \centering
 \includegraphics[width=\linewidth]{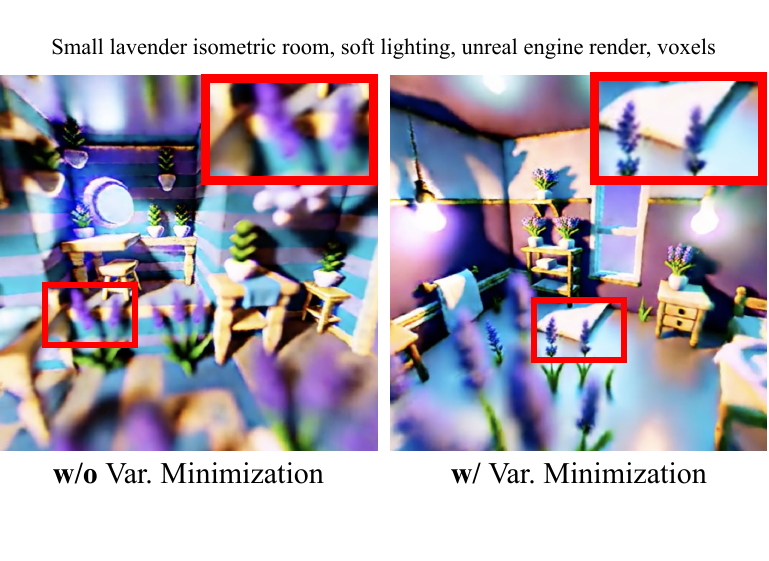}
 \vspace{-4em}
 \caption{\textbf{Ablation study on explicit variance minimization.} We study the effect of turning on/off the optimization step for $\Mat{\mu}$ with respect to loss Eq. \ref{eqn:var_min}.}
 \label{fig:ablation_min_var}
\vspace{-2em}
\end{figure}

To validate the effectiveness of our proposed components, we conduct ablation studies on whether or not to employ Eq.~\ref{eqn:var_min} to minimize second-order moment.
The alternative candidate is to fix $\Mat{\mu}$ as all-one vector during training.
As shown in Fig.~\ref{fig:ablation_min_var}, when explicitly minimizing the variance, we reach cleaner results with better high-frequency signals.
The results when optimizing without variance minimization, on the other hand, turned out to generate blurry geometry and noisy textures.
It is worth mentioning that excessive variance reduction may smoothen out some necessary details, especially in the background regions, as the left-hand side result contains more detailed background textures than the right-hand side one.

\subsection{Convergence Speed}
\begin{figure}[!t]
 \centering
 \includegraphics[width=\linewidth]{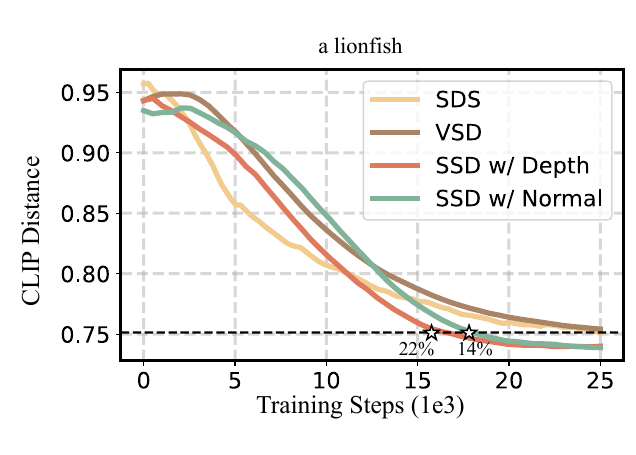}
 \vspace{-3em}
 \caption{\textbf{Convergence speed comparison.}  With the help of more stable gradient updates, SteinDreamer accelerates the training process by 14\%-22\%.}
 \label{fig:converge_speed}
\vspace{-1em}
\end{figure}

We also study the convergence speed of our methods as well as compared baselines. Specifically, we use the average CLIP distance \citep{xu2022neurallift} between the rendered images and the input text prompts as the quality metric. During the training process, we render the 3D assets into multi-view images every 1,000 training steps. In each training step, the diffusion model is inference twice through the classifier-free guidance, which is the same protocol in all compared methods. 
In Fig.~\ref{fig:converge_speed}, we profile the training steps needed for each approach to reach 0.75 CLIP distance as a desirable threshold.
We observe that the proposed SteinDreamer can effectively attain rapid and superior convergence, saving 14\%-22\% calls of diffusion models.
This means that lower variance in our distillation process can speed up convergence.
Our SteinDreamer utilizes fewer number of score function evaluations to achieve distilled 3D results that are more aligned with the text prompts.
Moreover, since SteinDreamer avoids inferencing and fine-tuning another diffusion model, each iteration of SSD is approximately 30\% faster than VSD (see Tab. \ref{tab:wall_clock} in Appendix \ref{sec:wall_time}).

\vspace{-2mm}
\section{Conclusion}
In this work, we present SteinDreamer, revealing a more general solution to reduce variance for score distillation. Our Stein Score Distillation (SSD) incorporates control variates through Stein identity, admitting arbitrary baseline functions conditioned
on all relevant variables with any guidance priors.
The experimental results suggest that SSD can effectively reduce the distillation variance and consistently improve visual quality for both object- and scene-level generations. We also showcase that SSD achieves faster and better convergence than existing methods. %

\section*{Acknowledgments}
P Wang sincerely appreciates insightful discussions with Zhaoyang Lv, Xiaoyu Xiang, Amit Kumar, Jinhui Xiong, and Varun Nagaraja.
P Wang also thanks Ruisi Cai for helping plot figures.
Any statements, opinions, findings, and conclusions or recommendations expressed in this material are those of the authors and do not necessarily reflect the views of their employers or the supporting entities.

\section*{Impact Statement}
This paper presents work whose goal is to advance the field of Machine Learning. However, there are many potential societal consequences inherent to generative models. For example, synthesizing harmful or offensive images or 3D assets.

\bibliography{example_paper}

\begin{thebibliography}{60}
\providecommand{\natexlab}[1]{#1}
\providecommand{\url}[1]{\texttt{#1}}
\expandafter\ifx\csname urlstyle\endcsname\relax
  \providecommand{\doi}[1]{doi: #1}\else
  \providecommand{\doi}{doi: \begingroup \urlstyle{rm}\Url}\fi

\bibitem[Cai et~al.(2020)Cai, Yang, Averbuch-Elor, Hao, Belongie, Snavely, and Hariharan]{cai2020learning}
Cai, R., Yang, G., Averbuch-Elor, H., Hao, Z., Belongie, S., Snavely, N., and Hariharan, B.
\newblock Learning gradient fields for shape generation.
\newblock In \emph{Computer Vision--ECCV 2020: 16th European Conference, Glasgow, UK, August 23--28, 2020, Proceedings, Part III 16}, pp.\  364--381. Springer, 2020.

\bibitem[Chan et~al.(2022)Chan, Lin, Chan, Nagano, Pan, De~Mello, Gallo, Guibas, Tremblay, Khamis, et~al.]{chan2022efficient}
Chan, E.~R., Lin, C.~Z., Chan, M.~A., Nagano, K., Pan, B., De~Mello, S., Gallo, O., Guibas, L.~J., Tremblay, J., Khamis, S., et~al.
\newblock Efficient geometry-aware 3d generative adversarial networks.
\newblock In \emph{Proceedings of the IEEE/CVF Conference on Computer Vision and Pattern Recognition}, pp.\  16123--16133, 2022.

\bibitem[Chen et~al.(2017)Chen, Zhu, and Song]{chen2017stochastic}
Chen, J., Zhu, J., and Song, L.
\newblock Stochastic training of graph convolutional networks with variance reduction.
\newblock \emph{arXiv preprint arXiv:1710.10568}, 2017.

\bibitem[Chen(1975)]{chen1975poisson}
Chen, L.~H.
\newblock Poisson approximation for dependent trials.
\newblock \emph{The Annals of Probability}, 3\penalty0 (3):\penalty0 534--545, 1975.

\bibitem[Chen et~al.(2023)Chen, Chen, Jiao, and Jia]{chen2023fantasia3d}
Chen, R., Chen, Y., Jiao, N., and Jia, K.
\newblock Fantasia3d: Disentangling geometry and appearance for high-quality text-to-3d content creation.
\newblock \emph{arXiv preprint arXiv:2303.13873}, 2023.

\bibitem[Davies et~al.(2004)Davies, Follana, Gray, Lepage, Mason, Nobes, Shigemitsu, Trottier, Wingate, Aubin, et~al.]{davies2004high}
Davies, C.~T., Follana, E., Gray, A., Lepage, G., Mason, Q., Nobes, M., Shigemitsu, J., Trottier, H., Wingate, M., Aubin, C., et~al.
\newblock High-precision lattice qcd confronts experiment.
\newblock \emph{Physical Review Letters}, 92\penalty0 (2):\penalty0 022001, 2004.

\bibitem[Detommaso et~al.(2018)Detommaso, Cui, Marzouk, Spantini, and Scheichl]{detommaso2018stein}
Detommaso, G., Cui, T., Marzouk, Y., Spantini, A., and Scheichl, R.
\newblock A stein variational newton method.
\newblock \emph{Advances in Neural Information Processing Systems}, 31, 2018.

\bibitem[Dhariwal \& Nichol(2021)Dhariwal and Nichol]{dhariwal2021diffusion}
Dhariwal, P. and Nichol, A.
\newblock Diffusion models beat gans on image synthesis.
\newblock \emph{Advances in neural information processing systems}, 34:\penalty0 8780--8794, 2021.

\bibitem[Garrigos \& Gower(2023)Garrigos and Gower]{garrigos2023handbook}
Garrigos, G. and Gower, R.~M.
\newblock Handbook of convergence theorems for (stochastic) gradient methods.
\newblock \emph{arXiv preprint arXiv:2301.11235}, 2023.

\bibitem[Goodfellow et~al.(2014)Goodfellow, Pouget-Abadie, Mirza, Xu, Warde-Farley, Ozair, Courville, and Bengio]{goodfellow2014generative}
Goodfellow, I., Pouget-Abadie, J., Mirza, M., Xu, B., Warde-Farley, D., Ozair, S., Courville, A., and Bengio, Y.
\newblock Generative adversarial nets.
\newblock \emph{Advances in neural information processing systems}, 27, 2014.

\bibitem[Gorham \& Mackey(2015)Gorham and Mackey]{gorham2015measuring}
Gorham, J. and Mackey, L.
\newblock Measuring sample quality with stein's method.
\newblock \emph{Advances in neural information processing systems}, 28, 2015.

\bibitem[Heusel et~al.(2017)Heusel, Ramsauer, Unterthiner, Nessler, and Hochreiter]{heusel2017gans}
Heusel, M., Ramsauer, H., Unterthiner, T., Nessler, B., and Hochreiter, S.
\newblock Gans trained by a two time-scale update rule converge to a local nash equilibrium.
\newblock \emph{Advances in neural information processing systems}, 30, 2017.

\bibitem[Ho et~al.(2020)Ho, Jain, and Abbeel]{ho2020denoising}
Ho, J., Jain, A., and Abbeel, P.
\newblock Denoising diffusion probabilistic models.
\newblock \emph{Advances in neural information processing systems}, 33:\penalty0 6840--6851, 2020.

\bibitem[H{\"o}llein et~al.(2023)H{\"o}llein, Cao, Owens, Johnson, and Nie{\ss}ner]{hollein2023text2room}
H{\"o}llein, L., Cao, A., Owens, A., Johnson, J., and Nie{\ss}ner, M.
\newblock Text2room: Extracting textured 3d meshes from 2d text-to-image models.
\newblock \emph{arXiv preprint arXiv:2303.11989}, 2023.

\bibitem[Hong et~al.(2023)Hong, Ahn, and Kim]{hong2023debiasing}
Hong, S., Ahn, D., and Kim, S.
\newblock Debiasing scores and prompts of 2d diffusion for robust text-to-3d generation.
\newblock \emph{arXiv preprint arXiv:2303.15413}, 2023.

\bibitem[Hu et~al.(2021)Hu, Shen, Wallis, Allen-Zhu, Li, Wang, Wang, and Chen]{hu2021lora}
Hu, E.~J., Shen, Y., Wallis, P., Allen-Zhu, Z., Li, Y., Wang, S., Wang, L., and Chen, W.
\newblock Lora: Low-rank adaptation of large language models.
\newblock \emph{arXiv preprint arXiv:2106.09685}, 2021.

\bibitem[Huang et~al.(2023)Huang, Wang, Shi, Qi, Zha, and Zhang]{huang2023dreamtime}
Huang, Y., Wang, J., Shi, Y., Qi, X., Zha, Z.-J., and Zhang, L.
\newblock Dreamtime: An improved optimization strategy for text-to-3d content creation.
\newblock \emph{arXiv preprint arXiv:2306.12422}, 2023.

\bibitem[Jain et~al.(2022)Jain, Mildenhall, Barron, Abbeel, and Poole]{jain2022zero}
Jain, A., Mildenhall, B., Barron, J.~T., Abbeel, P., and Poole, B.
\newblock Zero-shot text-guided object generation with dream fields.
\newblock In \emph{Proceedings of the IEEE/CVF Conference on Computer Vision and Pattern Recognition}, pp.\  867--876, 2022.

\bibitem[Jun \& Nichol(2023)Jun and Nichol]{jun2023shap}
Jun, H. and Nichol, A.
\newblock Shap-e: Generating conditional 3d implicit functions.
\newblock \emph{arXiv preprint arXiv:2305.02463}, 2023.

\bibitem[Kajiya(1986)]{kajiya1986rendering}
Kajiya, J.~T.
\newblock The rendering equation.
\newblock In \emph{Proceedings of the 13th annual conference on Computer graphics and interactive techniques}, pp.\  143--150, 1986.

\bibitem[Kim et~al.(2023)Kim, Lee, Choi, Jeong, Sohn, and Shin]{kim2023collaborative}
Kim, S., Lee, K., Choi, J.~S., Jeong, J., Sohn, K., and Shin, J.
\newblock Collaborative score distillation for consistent visual editing.
\newblock In \emph{Thirty-seventh Conference on Neural Information Processing Systems}, 2023.

\bibitem[Lin et~al.(2023)Lin, Gao, Tang, Takikawa, Zeng, Huang, Kreis, Fidler, Liu, and Lin]{lin2023magic3d}
Lin, C.-H., Gao, J., Tang, L., Takikawa, T., Zeng, X., Huang, X., Kreis, K., Fidler, S., Liu, M.-Y., and Lin, T.-Y.
\newblock Magic3d: High-resolution text-to-3d content creation.
\newblock In \emph{Proceedings of the IEEE/CVF Conference on Computer Vision and Pattern Recognition}, pp.\  300--309, 2023.

\bibitem[Liu et~al.(2017)Liu, Feng, Mao, Zhou, Peng, and Liu]{liu2017action}
Liu, H., Feng, Y., Mao, Y., Zhou, D., Peng, J., and Liu, Q.
\newblock Action-depedent control variates for policy optimization via stein's identity.
\newblock \emph{arXiv preprint arXiv:1710.11198}, 2017.

\bibitem[Liu(2017)]{liu2017stein}
Liu, Q.
\newblock Stein variational gradient descent as gradient flow.
\newblock \emph{Advances in neural information processing systems}, 30, 2017.

\bibitem[Liu \& Wang(2016)Liu and Wang]{liu2016stein}
Liu, Q. and Wang, D.
\newblock Stein variational gradient descent: A general purpose bayesian inference algorithm.
\newblock \emph{Advances in neural information processing systems}, 29, 2016.

\bibitem[Liu et~al.(2016)Liu, Lee, and Jordan]{liu2016kernelized}
Liu, Q., Lee, J., and Jordan, M.
\newblock A kernelized stein discrepancy for goodness-of-fit tests.
\newblock In \emph{International conference on machine learning}, pp.\  276--284. PMLR, 2016.

\bibitem[Liu et~al.(2023)Liu, Wu, Van~Hoorick, Tokmakov, Zakharov, and Vondrick]{liu2023zero}
Liu, R., Wu, R., Van~Hoorick, B., Tokmakov, P., Zakharov, S., and Vondrick, C.
\newblock Zero-1-to-3: Zero-shot one image to 3d object.
\newblock \emph{arXiv preprint arXiv:2303.11328}, 2023.

\bibitem[Max(1995)]{max1995optical}
Max, N.
\newblock Optical models for direct volume rendering.
\newblock \emph{IEEE Transactions on Visualization and Computer Graphics}, 1\penalty0 (2):\penalty0 99--108, 1995.

\bibitem[Metzer et~al.(2023)Metzer, Richardson, Patashnik, Giryes, and Cohen-Or]{metzer2023latent}
Metzer, G., Richardson, E., Patashnik, O., Giryes, R., and Cohen-Or, D.
\newblock Latent-nerf for shape-guided generation of 3d shapes and textures.
\newblock In \emph{Proceedings of the IEEE/CVF Conference on Computer Vision and Pattern Recognition}, pp.\  12663--12673, 2023.

\bibitem[Meyn(2008)]{meyn2008control}
Meyn, S.
\newblock \emph{Control techniques for complex networks}.
\newblock Cambridge University Press, 2008.

\bibitem[Mildenhall et~al.(2020)Mildenhall, Srinivasan, Tancik, Barron, Ramamoorthi, and Ng]{mildenhall2020nerf}
Mildenhall, B., Srinivasan, P.~P., Tancik, M., Barron, J.~T., Ramamoorthi, R., and Ng, R.
\newblock Nerf: Representing scenes as neural radiance fields for view synthesis.
\newblock In \emph{European conference on computer vision}, pp.\  405--421. Springer, 2020.

\bibitem[M{\"u}ller et~al.(2020)M{\"u}ller, Rousselle, Keller, and Nov{\'a}k]{muller2020neural}
M{\"u}ller, T., Rousselle, F., Keller, A., and Nov{\'a}k, J.
\newblock Neural control variates.
\newblock \emph{ACM Transactions on Graphics (TOG)}, 39\penalty0 (6):\penalty0 1--19, 2020.

\bibitem[M{\"u}ller et~al.(2022)M{\"u}ller, Evans, Schied, and Keller]{muller2022instant}
M{\"u}ller, T., Evans, A., Schied, C., and Keller, A.
\newblock Instant neural graphics primitives with a multiresolution hash encoding.
\newblock \emph{arXiv preprint arXiv:2201.05989}, 2022.

\bibitem[Nichol et~al.(2021)Nichol, Dhariwal, Ramesh, Shyam, Mishkin, McGrew, Sutskever, and Chen]{nichol2021glide}
Nichol, A., Dhariwal, P., Ramesh, A., Shyam, P., Mishkin, P., McGrew, B., Sutskever, I., and Chen, M.
\newblock Glide: Towards photorealistic image generation and editing with text-guided diffusion models.
\newblock \emph{arXiv preprint arXiv:2112.10741}, 2021.

\bibitem[Nichol et~al.(2022)Nichol, Jun, Dhariwal, Mishkin, and Chen]{nichol2022point}
Nichol, A., Jun, H., Dhariwal, P., Mishkin, P., and Chen, M.
\newblock Point-e: A system for generating 3d point clouds from complex prompts.
\newblock \emph{arXiv preprint arXiv:2212.08751}, 2022.

\bibitem[Oates et~al.(2017)Oates, Girolami, and Chopin]{oates2017control}
Oates, C.~J., Girolami, M., and Chopin, N.
\newblock Control functionals for monte carlo integration.
\newblock \emph{Journal of the Royal Statistical Society Series B: Statistical Methodology}, 79\penalty0 (3):\penalty0 695--718, 2017.

\bibitem[Poole et~al.(2022)Poole, Jain, Barron, and Mildenhall]{poole2022dreamfusion}
Poole, B., Jain, A., Barron, J.~T., and Mildenhall, B.
\newblock Dreamfusion: Text-to-3d using 2d diffusion.
\newblock \emph{arXiv preprint arXiv:2209.14988}, 2022.

\bibitem[Radford et~al.(2021)Radford, Kim, Hallacy, Ramesh, Goh, Agarwal, Sastry, Askell, Mishkin, Clark, et~al.]{radford2021learning}
Radford, A., Kim, J.~W., Hallacy, C., Ramesh, A., Goh, G., Agarwal, S., Sastry, G., Askell, A., Mishkin, P., Clark, J., et~al.
\newblock Learning transferable visual models from natural language supervision.
\newblock In \emph{International conference on machine learning}, pp.\  8748--8763. PMLR, 2021.

\bibitem[Ramesh et~al.(2021)Ramesh, Pavlov, Goh, Gray, Voss, Radford, Chen, and Sutskever]{ramesh2021zero}
Ramesh, A., Pavlov, M., Goh, G., Gray, S., Voss, C., Radford, A., Chen, M., and Sutskever, I.
\newblock Zero-shot text-to-image generation.
\newblock In \emph{International Conference on Machine Learning}, pp.\  8821--8831. PMLR, 2021.

\bibitem[Ramesh et~al.(2022)Ramesh, Dhariwal, Nichol, Chu, and Chen]{ramesh2022hierarchical}
Ramesh, A., Dhariwal, P., Nichol, A., Chu, C., and Chen, M.
\newblock Hierarchical text-conditional image generation with clip latents.
\newblock \emph{arXiv preprint arXiv:2204.06125}, 2022.

\bibitem[Ranftl et~al.(2020)Ranftl, Lasinger, Hafner, Schindler, and Koltun]{ranftl2020towards}
Ranftl, R., Lasinger, K., Hafner, D., Schindler, K., and Koltun, V.
\newblock Towards robust monocular depth estimation: Mixing datasets for zero-shot cross-dataset transfer.
\newblock \emph{IEEE transactions on pattern analysis and machine intelligence}, 44\penalty0 (3):\penalty0 1623--1637, 2020.

\bibitem[Ranftl et~al.(2021)Ranftl, Bochkovskiy, and Koltun]{ranftl2021vision}
Ranftl, R., Bochkovskiy, A., and Koltun, V.
\newblock Vision transformers for dense prediction.
\newblock In \emph{Proceedings of the IEEE/CVF international conference on computer vision}, pp.\  12179--12188, 2021.

\bibitem[Roeder et~al.(2017)Roeder, Wu, and Duvenaud]{roeder2017sticking}
Roeder, G., Wu, Y., and Duvenaud, D.~K.
\newblock Sticking the landing: Simple, lower-variance gradient estimators for variational inference.
\newblock \emph{Advances in Neural Information Processing Systems}, 30, 2017.

\bibitem[Shue et~al.(2022)Shue, Chan, Po, Ankner, Wu, and Wetzstein]{shue20223d}
Shue, J.~R., Chan, E.~R., Po, R., Ankner, Z., Wu, J., and Wetzstein, G.
\newblock 3d neural field generation using triplane diffusion.
\newblock \emph{arXiv preprint arXiv:2211.16677}, 2022.

\bibitem[Sohl-Dickstein et~al.(2015)Sohl-Dickstein, Weiss, Maheswaranathan, and Ganguli]{sohl2015deep}
Sohl-Dickstein, J., Weiss, E., Maheswaranathan, N., and Ganguli, S.
\newblock Deep unsupervised learning using nonequilibrium thermodynamics.
\newblock In \emph{International Conference on Machine Learning}, pp.\  2256--2265. PMLR, 2015.

\bibitem[Song \& Ermon(2019)Song and Ermon]{song2019generative}
Song, Y. and Ermon, S.
\newblock Generative modeling by estimating gradients of the data distribution.
\newblock \emph{Advances in neural information processing systems}, 32, 2019.

\bibitem[Song et~al.(2020)Song, Sohl-Dickstein, Kingma, Kumar, Ermon, and Poole]{song2020score}
Song, Y., Sohl-Dickstein, J., Kingma, D.~P., Kumar, A., Ermon, S., and Poole, B.
\newblock Score-based generative modeling through stochastic differential equations.
\newblock \emph{arXiv preprint arXiv:2011.13456}, 2020.

\bibitem[Stein(1972)]{stein1972bound}
Stein, C.
\newblock A bound for the error in the normal approximation to the distribution of a sum of dependent random variables.
\newblock In \emph{Proceedings of the Sixth Berkeley Symposium on Mathematical Statistics and Probability, Volume 2: Probability Theory}, volume~6, pp.\  583--603. University of California Press, 1972.

\bibitem[Sutton et~al.(1998)Sutton, Barto, et~al.]{sutton1998introduction}
Sutton, R.~S., Barto, A.~G., et~al.
\newblock \emph{Introduction to reinforcement learning}, volume 135.
\newblock MIT press Cambridge, 1998.

\bibitem[Sutton et~al.(1999)Sutton, McAllester, Singh, and Mansour]{sutton1999policy}
Sutton, R.~S., McAllester, D., Singh, S., and Mansour, Y.
\newblock Policy gradient methods for reinforcement learning with function approximation.
\newblock \emph{Advances in neural information processing systems}, 12, 1999.

\bibitem[Tsalicoglou et~al.(2023)Tsalicoglou, Manhardt, Tonioni, Niemeyer, and Tombari]{tsalicoglou2023textmesh}
Tsalicoglou, C., Manhardt, F., Tonioni, A., Niemeyer, M., and Tombari, F.
\newblock Textmesh: Generation of realistic 3d meshes from text prompts.
\newblock \emph{arXiv preprint arXiv:2304.12439}, 2023.

\bibitem[Wang et~al.(2023{\natexlab{a}})Wang, Du, Li, Yeh, and Shakhnarovich]{wang2023score}
Wang, H., Du, X., Li, J., Yeh, R.~A., and Shakhnarovich, G.
\newblock Score jacobian chaining: Lifting pretrained 2d diffusion models for 3d generation.
\newblock In \emph{Proceedings of the IEEE/CVF Conference on Computer Vision and Pattern Recognition}, pp.\  12619--12629, 2023{\natexlab{a}}.

\bibitem[Wang et~al.(2023{\natexlab{b}})Wang, Xu, Fan, Wang, Mohan, Iandola, Ranjan, Li, Liu, Wang, et~al.]{wang2023taming}
Wang, P., Xu, D., Fan, Z., Wang, D., Mohan, S., Iandola, F., Ranjan, R., Li, Y., Liu, Q., Wang, Z., et~al.
\newblock Taming mode collapse in score distillation for text-to-3d generation.
\newblock \emph{arXiv preprint arXiv:2401.00909}, 2023{\natexlab{b}}.

\bibitem[Wang et~al.(2023{\natexlab{c}})Wang, Lu, Wang, Bao, Li, Su, and Zhu]{wang2023prolificdreamer}
Wang, Z., Lu, C., Wang, Y., Bao, F., Li, C., Su, H., and Zhu, J.
\newblock Prolificdreamer: High-fidelity and diverse text-to-3d generation with variational score distillation.
\newblock \emph{arXiv preprint arXiv:2305.16213}, 2023{\natexlab{c}}.

\bibitem[Warburg et~al.(2023)Warburg, Weber, Tancik, Holynski, and Kanazawa]{warburg2023nerfbusters}
Warburg, F., Weber, E., Tancik, M., Holynski, A., and Kanazawa, A.
\newblock Nerfbusters: Removing ghostly artifacts from casually captured nerfs.
\newblock \emph{arXiv preprint arXiv:2304.10532}, 2023.

\bibitem[Williams(1992)]{williams1992simple}
Williams, R.~J.
\newblock Simple statistical gradient-following algorithms for connectionist reinforcement learning.
\newblock \emph{Machine learning}, 8:\penalty0 229--256, 1992.

\bibitem[Wu et~al.(2016)Wu, Zhang, Xue, Freeman, and Tenenbaum]{wu2016learning}
Wu, J., Zhang, C., Xue, T., Freeman, B., and Tenenbaum, J.
\newblock Learning a probabilistic latent space of object shapes via 3d generative-adversarial modeling.
\newblock \emph{Advances in neural information processing systems}, 29, 2016.

\bibitem[Xu et~al.(2022)Xu, Jiang, Wang, Fan, Wang, and Wang]{xu2022neurallift}
Xu, D., Jiang, Y., Wang, P., Fan, Z., Wang, Y., and Wang, Z.
\newblock Neurallift-360: Lifting an in-the-wild 2d photo to a 3d object with 360 $\{$$\backslash$deg$\}$ views.
\newblock \emph{arXiv preprint arXiv:2211.16431}, 2022.

\bibitem[Yang et~al.(2019)Yang, Huang, Hao, Liu, Belongie, and Hariharan]{yang2019pointflow}
Yang, G., Huang, X., Hao, Z., Liu, M.-Y., Belongie, S., and Hariharan, B.
\newblock Pointflow: 3d point cloud generation with continuous normalizing flows.
\newblock In \emph{Proceedings of the IEEE/CVF international conference on computer vision}, pp.\  4541--4550, 2019.

\bibitem[Zhang et~al.(2023)Zhang, Wang, Siarohin, Zhuang, Xu, Yang, Lin, Zhou, Tulyakov, and Lee]{zhang2023scenewiz3d}
Zhang, Q., Wang, C., Siarohin, A., Zhuang, P., Xu, Y., Yang, C., Lin, D., Zhou, B., Tulyakov, S., and Lee, H.-Y.
\newblock Scenewiz3d: Towards text-guided 3d scene composition.
\newblock \emph{arXiv preprint arXiv:2312.08885}, 2023.

\end{thebibliography}
\bibliographystyle{icml2024}

\newpage
\appendix
\onecolumn

\section{Deferred Derivations} \label{sec:app_proofs}

\paragraph{Gradient of KL divergence.}
Let $\Mat{\theta}$ parameterize the underlying 3D representation, such as NeRF \citep{mildenhall2020nerf}.
We intend to optimize $\Mat{\theta}$ such that each view matches the prior of 2D distribution. This can be formulated by minimizing the KL divergence below \footnote{Without loss of generality, we intend to omit coefficients $\omega(t)$ in all derivations for the sake of simplicity.}:
\begin{align} \label{eqn:kl_obj}
\min_{\Mat{\theta}} \mean_{t, \Mat{c} \sim p(\Mat{c})} \KL(q_t(\Mat{x}_t | \Mat{\theta}, \Mat{c}) \Vert p_t(\Mat{x}_t | \Mat{y})),
\end{align}
where $\Mat{c}$ is the camera pose sampled from a prior distribution, $\Mat{y}$ is the user-specified text prompt, and $q_t(\Mat{x}_t | \Mat{\theta}, \Mat{c}) = \gauss(\Mat{x}_t | \alpha_t g(\Mat{\theta}, \Mat{c}), \sigma_t^2 \Mat{I})$, where $g(\Mat{\theta}, \Mat{c})$ is a differentiable renderer that displays scene $\Mat{\theta}$ from the camera angle $\Mat{c}$.

To optimize Eq. \ref{eqn:kl_obj}, we take the gradient in terms of $\Mat{\theta}$ and derive the following update formula:
\begin{align}
&\nabla_{\Mat{\theta}} \mean_{t, \Mat{c}} \KL(q_t(\Mat{x}_t | \Mat{\theta}, \Mat{c}) \Vert p_t(\Mat{x}_t | \Mat{y})) = \mathbb{E}_{t, \Mat{c}} \nabla_{\theta} \KL(q_t(\Mat{x}_t | \Mat{\theta}, \Mat{c}) \Vert p_t(\Mat{x}_t | \Mat{y})) \\
&= \mean_{t, \Mat{c}} \nabla_{\Mat{\theta}} \mean_{\Mat{x}_t \sim q_t(\Mat{x}_t | \Mat{\theta}, \Mat{c})} \left[ \log\frac{q_t(\Mat{x}_t | \Mat{\theta}, \Mat{c})}{p_t(\Mat{x}_t | \Mat{y})} \right] \\
&= \mean_{t, \Mat{c}, \Mat{\epsilon} \sim \gauss(\Mat{0}, \sigma_t^2 \Mat{I})} \left[ \underbrace{\nabla_{\Mat{\theta}} \log q_t(\alpha_t g(\Mat{\theta}, \Mat{c}) + \Mat{\epsilon} | \Mat{\theta}, \Mat{c})}_{(a)} - \underbrace{\nabla_{\Mat{\theta}} \log p_t(\alpha_t g(\Mat{\theta}, \Mat{c}) + \Mat{\epsilon} | \Mat{y})}_{(b)} \right]
\end{align}

We notice that $q_t(\alpha_t g(\Mat{\theta}, \Mat{c}) + \Mat{\epsilon} | \Mat{\theta}, \Mat{c}) = \gauss(\Mat{\epsilon} | \Mat{0}, \sigma_t^2 \Mat{I})$, which is independent of $\Mat{\theta}$. Thus $(a) = \Mat{0}$.
For term (b), we have:
\begin{align}
\nabla_{\Mat{\theta}} \log p_t(\alpha_t g(\Mat{\theta}, \Mat{c}) +\Mat{\epsilon} | \Mat{y}) = \alpha_t \frac{\partial g(\Mat{\theta}, \Mat{c})}{\partial \Mat{\theta}} \nabla \log p_t(\alpha_t g(\Mat{\theta}, \Mat{c}) + \Mat{\epsilon} | \Mat{y}).  
\end{align}
Therefore, $\Mat{\theta}$ should be iteratively updated by:
\begin{align} \label{eqn:kl_grad}
\mean_{t, \Mat{c}, \Mat{\epsilon}} \left[ \alpha_t \frac{\partial g(\Mat{\theta}, \Mat{c})}{\partial \Mat{\theta}} \nabla \log p_t(\alpha_t g(\Mat{\theta}, \Mat{c}) + \Mat{\epsilon} | \Mat{y})\right] = \mean_{t, \Mat{c},\Mat{x}_t \sim q_t(\Mat{x}_t | \Mat{\theta}, \Mat{c})} \left[ \alpha_t \frac{\partial g(\Mat{\theta}, \Mat{c})}{\partial \Mat{\theta}} \nabla \log p_t(\Mat{x}_t | \Mat{y})\right]
\end{align}

\paragraph{SDS equals to the gradient of KL.}
By the following derivation, we demonstrate that SDS essentially minimizes the KL divergence: $\Mat{\Delta}_{SDS} = \nabla_{\Mat{\theta}} \mean_{t, \Mat{c}} \KL(q_t(\Mat{x}_t | \Mat{\theta}, \Mat{c}) \Vert p_t(\Mat{x}_t | \Mat{y}))$:
\begin{align}
& \mean_{t, \Mat{c}, \Mat{x}_t \sim q_t(\Mat{x}_t | \Mat{\theta}, \Mat{c})} \left[ \frac{\partial g(\Mat{\theta}, \Mat{c})}{\partial \Mat{\theta}} \left(\nabla \log p_t(\Mat{x}_t | \Mat{y}) - \Mat{\epsilon}\right) \right] \\
&= \mean_{t, \Mat{c}, \Mat{x}_t \sim q_t(\Mat{x}_t | \Mat{\theta}, \Mat{c})} \left[ \alpha_t \frac{\partial g(\Mat{\theta}, \Mat{c})}{\partial \Mat{\theta}} \nabla \log p_t(\Mat{x}_t | \Mat{y}) \right] - \underbrace{\mean_{t, \Mat{c}, \Mat{\epsilon} \sim \gauss(\Mat{0}, \sigma_t^2 \Mat{I})} \left[ \alpha_t \frac{\partial g(\Mat{\theta}, \Mat{c})}{\partial \Mat{\theta}} \Mat{\epsilon} \right]}_{=\Mat{0}}.
\end{align}

\paragraph{VSD equals to the gradient of KL.}
We show that VSD also equals to the gradient of KL $\Mat{\Delta}_{VSD} = \nabla_{\Mat{\theta}} \mean_{t, \Mat{c}} \KL(q_t(\Mat{x}_t | \Mat{\theta}, \Mat{c}) \Vert p_t(\Mat{x}_t | \Mat{y}))$ due to the simple fact that the first-order of score equals to zero:
\begin{align}
& \mean_{t, \Mat{c}, \Mat{x}_t \sim q_t(\Mat{x}_t | \Mat{\theta}, \Mat{c})} \left[ \alpha_t \frac{\partial g(\Mat{\theta}, \Mat{c})}{\partial \Mat{\theta}} \nabla\log q_t(\Mat{x}_t | \Mat{\theta}, \Mat{c}) \right] = \mean_{t, \Mat{c}, \Mat{x}_t \sim q_t(\Mat{x}_t | \Mat{\theta}, \Mat{c})} \left[ \nabla_{\Mat{\theta}} \log q_t(\Mat{x}_t | \Mat{\theta}, \Mat{c}) \right] \\
&= \mean_{t, \Mat{c}} \left[ \int \frac{\nabla_{\Mat{\theta}} q_t(\Mat{x}_t | \Mat{\theta}, \Mat{c})}{q_t(\Mat{x}_t | \Mat{\theta}, \Mat{c})} q_t(\Mat{x}_t | \Mat{\theta}, \Mat{c}) d\Mat{x}_t \right] \\
&= \mean_{t, \Mat{c}} \left[\nabla_{\Mat{\theta}} \int q_t(\Mat{x}_t | \Mat{\theta}, \Mat{c}) d\Mat{x}_t \right] = \Mat{0}.
\end{align}

\paragraph{Control Variate for SSD.}
Due to Stein's identity, the following is constantly zero:
\begin{align}
\mean_{\Mat{x}_t \sim q_t(\Mat{x}_t | \Mat{\theta}, \Mat{c})} \left[ \nabla \log q_t(\Mat{x}_t | \Mat{\theta}, \Mat{c}) \phi(t, \Mat{\theta}, \Mat{x}_t, \Mat{c}) + \nabla_{\Mat{x}_t} \phi(t, \Mat{\theta}, \Mat{x}_t, \Mat{c}) \right] = \Mat{0}.
\end{align}
Plug into Eq. \ref{eqn:kl_grad}, we can obtain:
\begin{align}
& \mean_{t, \Mat{c},\Mat{x}_t \sim q_t(\Mat{x}_t | \Mat{\theta}, \Mat{c})} \left[ \alpha_t \frac{\partial g(\Mat{\theta}, \Mat{c})}{\partial \Mat{\theta}} \nabla \log p_t(\Mat{x}_t | \Mat{y})\right] \\
&= \mean_{t, \Mat{c}} \left[\frac{\partial g(\Mat{\theta}, \Mat{c})}{\partial \Mat{\theta}} \mean_{\Mat{x}_t \sim q_t(\Mat{x}_t | \Mat{\theta}, \Mat{c})} \left[ \nabla \log p_t(\Mat{x}_t | \Mat{y}) + \nabla \log q_t(\Mat{x}_t | \Mat{\theta}, \Mat{c}) \phi(t, \Mat{\theta}, \Mat{x}_t, \Mat{c}) + \nabla_{\Mat{x}_t} \phi(t, \Mat{\theta}, \Mat{x}_t, \Mat{c}) \right]\right] \\
&= \mean_{t, \Mat{c}} \left[\frac{\partial g(\Mat{\theta}, \Mat{c})}{\partial \Mat{\theta}} \mean_{\Mat{x}_t \sim q_t(\Mat{x}_t | \Mat{\theta}, \Mat{c})} \left[ \nabla \log p_t(\Mat{x}_t | \Mat{y}) + \Mat{\epsilon} \phi(t, \Mat{\theta}, \Mat{x}_t, \Mat{c}) + \nabla_{\Mat{x}_t} \phi(t, \Mat{\theta}, \Mat{x}_t, \Mat{c}) \right]\right] \label{eqn:score_to_noise} \\
&= \mean_{t, \Mat{c}, \Mat{\epsilon}} \left[ \frac{\partial g(\Mat{\theta}, \Mat{c})}{\partial \Mat{\theta}} (\nabla \log p_t(\Mat{x}_t | \Mat{y}) + \Mat{\epsilon} \phi(t, \Mat{\theta}, \Mat{x}_t, \Mat{c}) + \nabla_{\Mat{x}_t} \phi(t, \Mat{\theta}, \Mat{x}_t, \Mat{c})) \right],
\end{align}
where Eq. \ref{eqn:score_to_noise} can be derived by noticing $q_t(\Mat{x}_t | \Mat{\theta}, \Mat{c})$ follows from a Gaussian distribution.

\section{Deferred Discussion}
\label{sec:app_discuss}

In this section, we continue our discussion from Sec. \ref{sec:discuss}.

\paragraph{How does baseline function reduce variance?}
The baseline function $\phi$ can be regarded as a guidance introduced into the distillation process.
We contend that control variates when equipped with pre-trained models incorporating appropriate 2D/3D prior knowledge, are likely to exhibit a higher correlation with the score function.
Intuitively, enforcing priors and constraints on the gradient space can also stabilize the training process by regularizing the optimization trajectory. 
Therefore, in our empirical design, the inclusion of geometric information expressed by a pre-trained MiDAS estimator is expected to result in superior variance reduction compared to SSD and VSD.

\paragraph{Comparison with VSD.}
In VSD, the adopted control variate $\nabla \log q_t(\Mat{x}_t | \Mat{c})$ is fine-tuned based on a pre-trained score function using LoRA \citep{hu2021lora}.
However, this approach presents two primary drawbacks: 
1) The trained control variate may not fully converge to the desired score function, potentially resulting in non-zero mean and biased gradient estimations. 2) Fine-tuning another large diffusion model also significantly increases the computation expenses.
Our SSD effectively circumvents these two limitations.
Firstly, the control variate in SSD is provably zero-mean, as per Stein's identity. Additionally, the computational cost associated with differentiating the frozen $\phi$ and optimizing the weights $\Mat{u}$ remains manageable.
We verify the computational efficiency of SSD in Appendix \ref{sec:wall_time}.

\paragraph{Other baseline functions.}
It is noteworthy that baseline functions other than depth/normal predictors are also applicable.
As we discussed above, choosing the right baseline functions can implicitly incorporate desirable prior information.
Here we provide some tentative options for future exploration.
A foreground-background segmenter coupled with a classification loss can be useful to mitigate the artifacts of missing parts in generated 3D objects.
A discriminator from a pre-trained GAN \citep{goodfellow2014generative} associated with the discriminative loss can be utilized to implement the baseline function to improve the fidelity of each view.
Similarly, CLIP loss \citep{jain2022zero} inducted baseline function might help increase relevance with specified text. 
Multi-view prior such as Zero123 \citep{liu2023zero} can be also introduced by sampling another view as a function of $\Mat{\theta}$ and comparing it with the current view $\Mat{x}_t$.
Notably, our method supports freely combining all these aforementioned baseline functions.

\section{Additional Experiments}

\subsection{Implementation Details}
\label{sec:app_expr_details}

\paragraph{Image Sampling.}
In our 2D experiment in Fig. \ref{fig:var_vsd_sds}, one could simply regard $g(\Mat{\theta}) = \Mat{\theta}$ where $\Mat{\theta}$ is an image representation.
We train all algorithms for 5k steps with learning rate 1e-2. For VSD, we update its LoRA with learning rate 1e-4. 
When sampling with SSD, we utilize CLIP guidance to construct the baseline function, akin to \citet{xu2022neurallift}.
In particular, baseline function is defined as $\phi(t, \Mat{x}_t, \Mat{\theta}, \Mat{y}) = -\langle \operatorname{CLIP}_{image}(\Mat{\theta}),  \operatorname{CLIP}_{text}(\Mat{y}) \rangle$.

\paragraph{3D Generation.}

All of baselines are implemented based on the \texttt{threestudio} framework.
For fairness, we only compare the results yielded in the coarse generation stage for object generation without geometry refinement specified in ProlificDreamer.
We employ hash-encoded NeRF \citep{muller2022instant} as the underlying 3D representation, and disentangle representations for foreground and background separately.
All scenes are trained for 25k steps with a learning rate of 1e-2.
At each iteration, we randomly sample one view for supervision.
We progressively increase rendering resolution from 64$\times$64 resolution to 256$\times$256 resolution after 5k steps.
View-dependent prompting is enabled to alleviate Janus problem.
Other hyperparameters are kept consistent with the default values.
In our implementation of SteinDreamer, the depth estimator is operated on images decoded from the latent space.
We further scale the baseline function by a coefficient 1e-2.
The pre-trained MiDAS is based on a hybrid architecture of transformer and ResNet \citep{ranftl2021vision}.
We convert the estimated inverse depth to normal by directly taking spatial derivatives and normalization. Such operation is equivalent to the standard computation via normalizing spatial derivatives of non-inverse depth.
The rendered normal is analytically computed as the gradient of the density field.
Additionally, we reweight the pre-average loss map via the alpha mask rendered from NeRF.
The weighting coefficients $\Mat{\mu}$ are initialized with an all-one vector.

\subsection{Wall-Clock Time Benchmarking}
\label{sec:wall_time}

In addition to faster convergence, we also test per-iteration wall-clock time for all methods.
Results are listed in Tab. \ref{tab:wall_clock}.
The reported numbers are obtained by averaging the running time of corresponding methods with six prompts in Tab. \ref{tab:obj_quantitative} for 10k iterations \textit{on the same device}.
In summary, SteinDreamer exhibits comparable per-iteration speed to SDS while significantly outperforming VSD in terms of speed. The trainable component $\Mat{\mu}$ in SSD comprises only thousands of parameters, which minimally increases computational overhead and becomes much more efficient than tuning a LoRA in VSD. Notably, given that SSD can reach comparable visual quality in fewer steps, SteinDreamer achieves significant time savings for 3D score distillation.

\begin{table}[!h]
\centering
\begin{tabular}{lrrr}
\toprule
 Methods &   Sec. / Iter. \\
\hline
SDS \citep{poole2022dreamfusion}      &  1.063 $\pm$ 0.002  \\
VSD \citep{wang2023prolificdreamer}     &  1.550 $\pm$ 0.004  \\
SSD w/ Depth (Ours)  &  1.093 $\pm$ 0.005 \\
SSD w/ Normal (Ours) &    1.087 $\pm$ 0.004 \\
\bottomrule
\end{tabular}
\caption{\textbf{Benchmarking wall-clock time}. We test wall-clock time (seconds per iteration) for all considered methods. }
\label{tab:wall_clock}
\end{table}

\begin{figure}[b]
    \centering
    \includegraphics[width=0.9\textwidth]{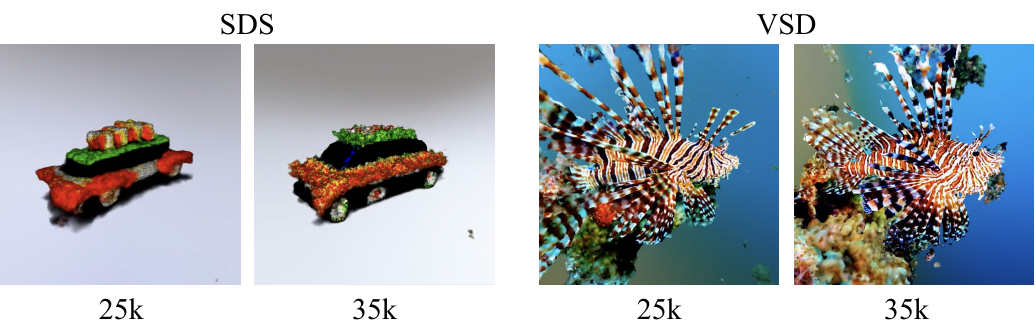}
    \vspace{-1em}
    \caption{\small \textbf{Longer Training Results.} We train high-variance score distillation approaches SDS and VSD for extra 10k steps. Prompts: `` car made out of sush'' for SDS and ``a lionfish'' for VSD}
    \label{fig:longer_training}
\end{figure}

\subsection{Longer Training for Baselines}

A naive solution to achieve better convergence with high-variance gradient descent is to increase training steps.
We test this hypothesis in this section by training SDS and VSD on two scenes with 10k more steps.
Qualitative results are presented in Fig. \ref{fig:longer_training}.
We notice that longer training time cannot guarantee better convergence.
We also quantitatively find that more optimization steps have negligible influence on the final CLIP scores, which float between 0.84 \textasciitilde 0.86 for the prompt `` car made out of sush'' and 0.74 \textasciitilde 0.75 for the prompt ``a lionfish''.

\begin{figure*}[h]
    \centering
    \includegraphics[width=\textwidth]{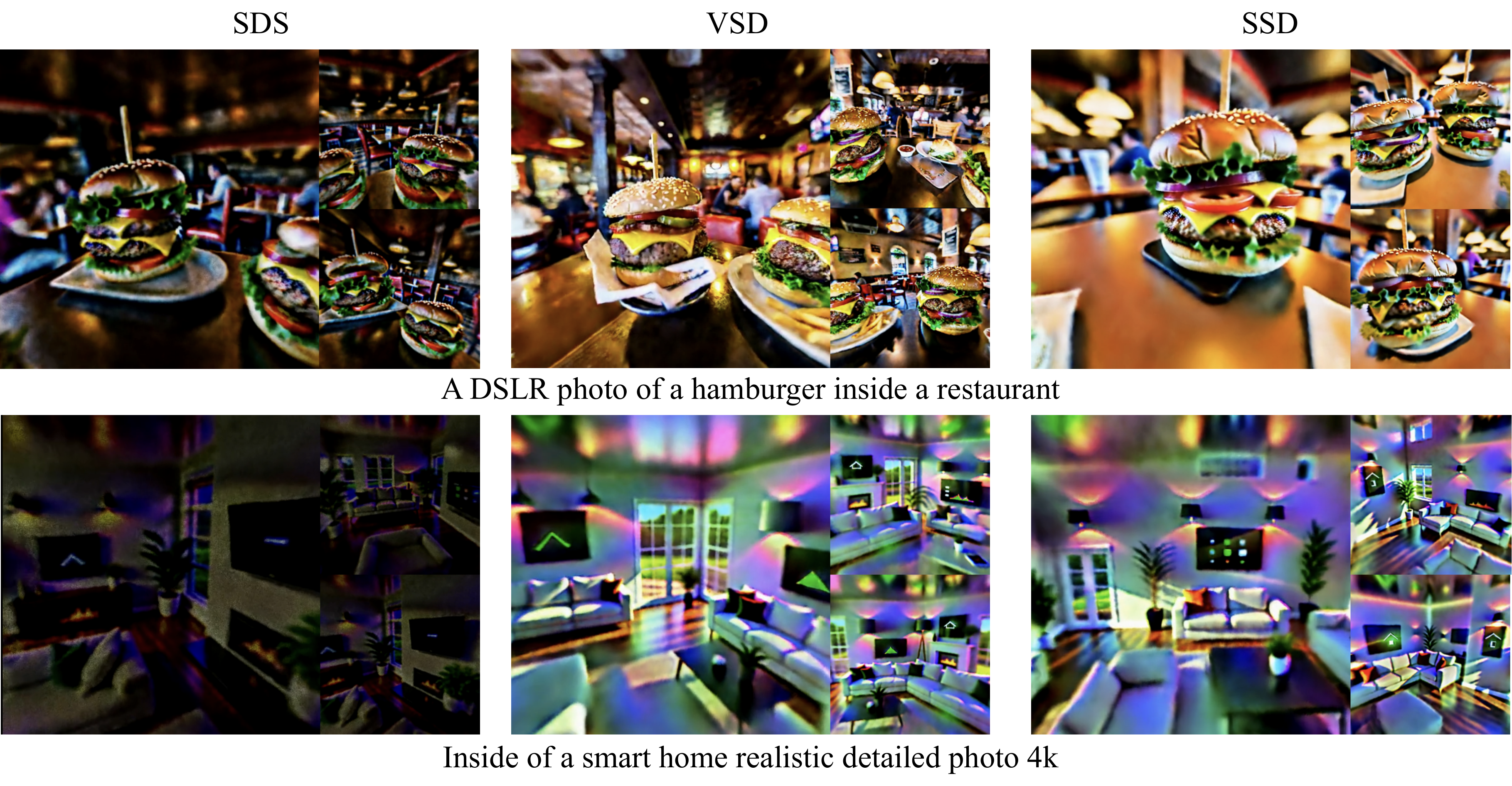}
    \caption{\textbf{Scene-level qualitative comparisons between DreamFusion, ProlificDreamer, and SteinDreamer w/ depth estimator.} Compared to existing methods,  SteinDreamer presents more realistic textures with better details.}
    \label{fig:scene_res_steindepth}
\end{figure*}

In optimization theory, variance plays a crucial role in determining the convergence rate of SGD algorithms \citep{garrigos2023handbook}
With a finite number of optimization steps and a standard learning rate, maintaining low variance is pivotal to ensure convergence.
Training with noisy gradients introduces high instability, potentially resulting in a suboptimal solution or even divergence, as illustrated in Fig. \ref{fig:scene_res_steindepth}.

\subsection{More Quantitative Results}
\label{sec:app_quant_res}

In addition to the overall quantitative comparison in Sec. \ref{sec:expr_res}, we also provide a breakdown table of the CLIP distance for every demonstrated object in this paper.
The numerical evaluation of these results is reported in Tab. \ref{tab:obj_quantitative}.
Our observations indicate that SteinDreamer consistently outperforms all other methods, which improves CLIP score by \textasciitilde0.5 over ProlificDreamer. This superior result suggests our flexible control variates are more effective than the one adopted by ProlificDreamer.

\begin{table}[!h]
\centering
\begin{tabular}{lrrr}
\toprule
 Methods              &   ``blue tulip'' &   ``sushi car'' &   ``lionfish'' \\
\hline
 SDS \citep{poole2022dreamfusion}      &   0.777 &       0.862 &      0.751 \\
 VSD \citep{wang2023prolificdreamer}     &   0.751 &       0.835 &      0.749 \\
 SSD w/ Depth (Ours)  &   \textbf{0.734} &      \textbf{0.754} &      \textbf{0.735} \\
\hline
               &   ``cheesecake castle'' &   ``dragon toy'' &   ``dog statue'' \\
\hline
 SDS \citep{poole2022dreamfusion}     &               0.902 &        0.904 &              0.789 \\
 VSD \citep{wang2023prolificdreamer}    &               0.843 &        0.852 &              0.775 \\
 SSD w/ Normal (Ours) &               \textbf{0.794} &        \textbf{0.806} &              \textbf{0.751} \\
\bottomrule
\end{tabular}
\caption{\textbf{Breakdown table.} We compare the CLIP distance ($\downarrow$ the lower the better) of demonstrated results among different approaches. Best results are marked in \textbf{bold} font. Prompts: ``blue tulip'' is short for ``a blue tulip'', ``sushi car'' for ``a car made out of sushi'', ``lionfish'' for ``a lionfish'', ``cheesecake castle'' for ``a Matte painting of a castle made of cheesecake surrounded by a moat made of ice cream'', ``dragon toy'' for ``a plush dragon toy'', and ``dog statue'' for ``michelangelo style statue of dog reading news on a cellphone''. }
\label{tab:obj_quantitative}
\end{table}

We also provide a list that shows other prompts involved into quantitative evaluation for object generation.
\begin{table}[h]
\centering
\begin{tabular}{l}
\toprule
\hline
a pineapple \\
a 3D model of an adorable cottage with a thatched roof \\
an elephant skull \\
a plate piled high with chocolate chip cookies \\
michelangelo style statue of dog reading news on a cellphone \\
a chimpanzee dressed like Henry VIII king of England \\
a delicious croissant \\
a sliced loaf of fresh bread \\
a small saguaro cactus planted in a clay pot \\
a blue tulip \\
a plate of fried chicken and waffles with maple syrup on them \\
a cauldron full of gold coins \\
a rabbit, animated movie character, high detail 3d model \\
a lionfish \\
a car made out of sushi \\
a DSLR photo of an imperial state crown of England \\
a rotary telephone carved out of wood \\
a marble bust of a mouse \\
a typewriter \\
a Matte painting of a castle made of cheesecake surrounded by a moat made of ice cream \\
a plush dragon toy \\
a praying mantis wearing roller \\
\bottomrule
\end{tabular}
\caption{\textbf{A list of text prompts.} All text prompts are collected from \citet{wang2023prolificdreamer}.}
\label{tab:prompts}
\end{table}

\subsection{More Qualitative Results} \label{sec:more_scene_res}

\begin{figure*}[h]
    \centering
     \includegraphics[width=\textwidth]{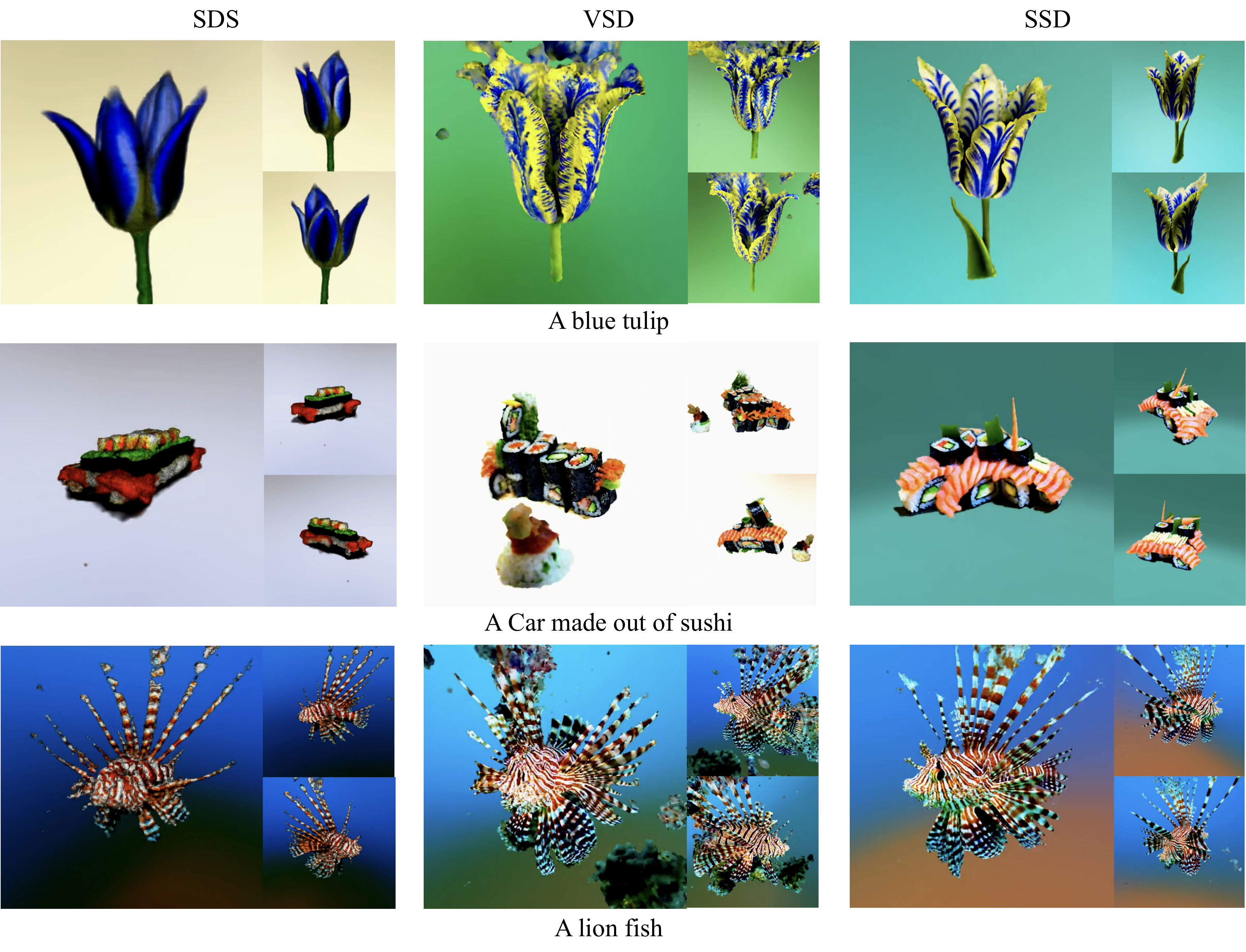}
     \vspace{-1em}
    \caption{\textbf{Object-level qualitative comparisons.} Compared to existing methods, our SteinDreamer w/ depth estimator delivers smoother geometry, more detailed texture, and fewer floater artifacts.}
    \label{fig:obj_res_steindepth}
\end{figure*}

\begin{figure*}[h]
    \centering
    \includegraphics[width=\textwidth]{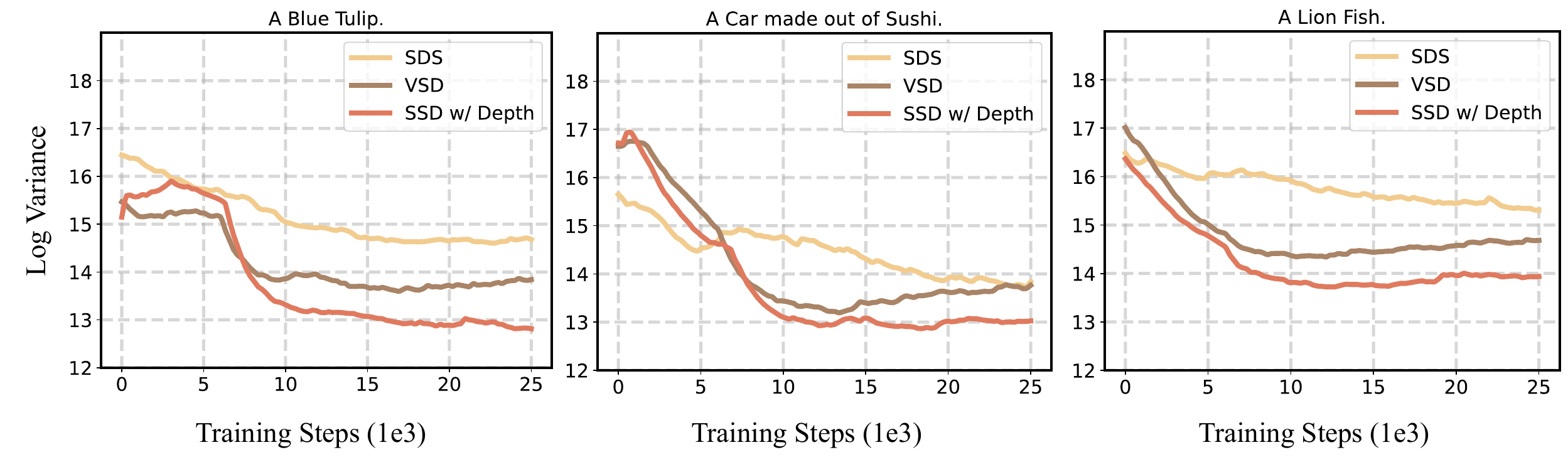}
    \caption{\small \textbf{Variance comparison of $\Mat{\Delta}_{SDS}$, $\Mat{\Delta}_{VSD}$, and $\Mat{\Delta}_{SSD}$ with depth estimator.}  We visualize how the variance of the investigated three methods for every 1,000 steps. The variance decays as the training converges while $\Mat{\Delta}_{SSD}$ consistently achieves lower variance throughout the whole process. }
    \label{fig:var_vsd_sds_ssd_depth}
    \vspace{-1em}
\end{figure*}

We demonstrate deferred results' of SteinDreamer with depth estimator in Fig. \ref{fig:obj_res_steindepth} and Fig. \ref{fig:obj_res_steinnormal}.
Consistent with our observation in Sec. \ref{sec:expr_res}, our method yields smooth and consistent renderings.
We refer interested readers to our supplementary materials for video demos.
Moreover, we visualize the comparison on variance during object generation in Fig. \ref{fig:var_vsd_sds_ssd_depth}.
Similarly, it validates that score distillation's performance is highly correlated with variance of stochastic gradients.

\end{document}